\documentclass[preprint,review,3p]{elsarticle}
\usepackage{color}
\usepackage{graphicx}
\usepackage{subcaption}
\usepackage{algorithm}
\usepackage{bm}
\usepackage[colorlinks]{hyperref}
\usepackage{amssymb}
\usepackage{amsthm}
\usepackage{amsmath}
\usepackage{booktabs}
\usepackage{url}
\usepackage{scrextend}
\usepackage{epstopdf}
\usepackage{float}
\usepackage[perpage]{footmisc}
\usepackage{tcolorbox}
\usepackage{lineno}


\biboptions{square}
\journal{arXiv}

\begin{document}

\begin{frontmatter}

\title{Probabilistic Performance-Pattern Decomposition (PPPD): analysis framework and applications to stochastic mechanical systems}
 \author[1]{Ziqi Wang}
 \author[2]{Marco Broccardo}
 \author[3]{Junho Song}
 \address[1]{Earthquake Engineering Research \& Test Center, Guangzhou University, China}
 \address[2]{Swiss Seismological Service, SED, ETH Z\"urich, Switzerland}
 \address[3]{Department of Civil and Environmental Engineering, Seoul National University, South Korea}
 \renewcommand{\thefootnote}{\alph{footnote}}
\footnotetext[1]{Corresponding author: \href{mailto:ziqidwang@yahoo.com}{ziqidwang@yahoo.com}}
\footnotetext[2]{Corresponding author: \href{mailto:ocramod@gmail.com}{bromarco@ethz.ch}}
\footnotetext[3]{Corresponding author: \href{mailto:junhosong@snu.ac.kr}{junhosong@snu.ac.kr}}

\begin{abstract}
Since the early 1900s, numerous research efforts have been devoted to developing quantitative solutions to stochastic mechanical systems. In general, the problem is perceived as solved when a complete or partial probabilistic description on the quantity of interest (QoI) is determined. However, in the presence of complex system behavior, there is a critical need to go beyond mere probabilistic descriptions. In fact, to gain a full understanding of the system,  it is crucial to extract physical characterizations from the probabilistic structure of the QoI, especially when the QoI solution is obtained in a data-driven fashion. Motivated by this perspective, the paper proposes a framework to obtain structuralized characterizations on behaviors of stochastic systems. The framework is named Probabilistic Performance-Pattern Decomposition (PPPD). PPPD analysis aims to decompose complex response behaviors, conditional to a prescribed performance state, into meaningful patterns in the space of system responses, and to investigate how the patterns are triggered in the space of basic random variables. To illustrate the application of PPPD, the paper studies three numerical examples: 1) an illustrative example with hypothetical stochastic processes input and output; 2) a stochastic Lorenz system with periodic as well as chaotic behaviors; and 3) a simplified shear-building model subjected to a stochastic ground motion excitation. 
\end{abstract}

\begin{keyword}
Autoencoder \sep clustering \sep diffusion map \sep  manifold learning \sep Monte Carlo simulation \sep pattern recognition \sep stochastic dynamics \sep uncertainty quantification
\end{keyword}

\end{frontmatter}

\renewcommand{\thefootnote}{\arabic{footnote}}

\section{Introduction}\label{introduction}
\noindent
The study of classical mechanics in the presence of uncertainties has become a crucial research topic in engineering, drawing a growing number of studies and applications. The topic finds its root in the study of Brownian motions in the early 1900s, following the pioneering works of Smoluchowski \cite{Smoluchowski}, Einstein \cite{Einstein}, Langevin \cite{Langevin}, It{\^o} \cite{Ito}, and Stratonovich \cite{stratonovich1966a}. Over time, the scope and depth of the subject have grown to include theoretical studies on stochastic differential equations \cite{SDE}\cite{higham2001an}\cite{grigoriu_2005}, stochastic dynamics \cite{RevComptStochMech}\cite{soong_grigoriu_1993}\cite{roberts_spanos_2003}\cite{li_chen_2009}, and risk/reliability theory \cite{ditlevsen1996str}\cite{ReliabilitySIAM}\cite{freudenthal1964analysis} to cite a few of them. In parallel, the number of engineering applications has also flourished, and an incomplete list includes \cite{ELLINGWOOD2001251}\cite{frangopol2003life-cycle}\cite{moller2008engineering}\cite{kougioumtzoglou2012response}.

Similar to deterministic mechanical systems, the starting point of the analysis of stochastic mechanical systems is  the equation of motion (typically in a form of stochastic differential equation). Unlike deterministic systems, the state of a stochastic mechanical system is a (not necessarily finite) set of random variables and the final output is a complete or partial probabilistic characterization on Quantities of Interest (QoIs).  In most cases, even approximate solutions are difficult to obtain \cite{SDE}\cite{higham2001an}. An alternative and popular approach is to recast the equation of motion into an equation of probability density function (Smoluchowski/Fokker–Planck equation \cite{jordan1998the}), or an equation of statistical moment (moment closure \cite{Crandall1985Non}). Unfortunately, these equations are also difficult to solve for generic multi-degree-of-freedom systems. Alternatively, in the recent years, with the rising and formalization of Uncertainty Quantification (UQ) as a new pillar of Engineering Science \cite{oden2010computer} \cite{ElishakoffUncertainty}\cite{roy2011a}\cite{OUQ}, the solution of stochastic mechanical systems can be cast as classical forward UQ problem. In particular, non-intrusive UQ methods \cite{Couaillier2019}\cite{Eldred09recentadvances} are appealing because they decouple the deterministic solution of the governing equations  (considered in a black-box fashion) from the statistical analysis on the input-QoI relationship. In engineering applications, this strategy is particularly convenient given the vast legacy of complex computational codes, which cannot be customized intrusively for UQ analysis. Given this premises, and with the advent of high-performance computing \cite{yasar2001new}, it is no surprise that Monte Carlo simulation methods \cite{MCS}\cite{neal2001annealed}\cite{au2007application} and metamodeling \cite{PCE}\cite{Echard2011AK}\cite{Sudret2012Meta} are becoming the way-to-go for UQ forward analysis and, consequently, for solving stochastic mechanical problems.

The use of classical non-intrusive UQ methods has been a remarkable advancement for the solution of stochastic mechanical problems. However, it has also stimulated an undesirable consequence: the problem of interest is perceived as solved once the probabilistic characterization of the QoIs is obtained. In fact, the classical UQ analysis and the following decision-making process are merely based on the statistics of the QoIs, loosing \textit{de facto} the physical information of the mechanical problem (hidden within the black-box solver). There is a critical missing link in this context, that is extracting physical information and patterns from the probabilistic characterization of the QoIs. This is crucial, especially in the presence of multi-degree-of-freedom systems with complex behavior. Therefore, this study aims to fill this research gap by defining a formal framework for extracting a global physical characterization from the probabilistic structure of the QoIs. Within the non-intrusive UQ perimeter, the ultimate goal is to promote a physics informed decision process, which focuses not only on the statistics of the QoIs but—more importantly—on the physical patterns that triggered such probabilistic representation.

We named the proposed framework Probabilistic Performance-Pattern Decomposition (PPPD). The term performance-pattern is adopted since we consider a behavior domain defined by performance state. Specifically, the paper develops methods to study behavior patterns of a complex stochastic system, and to identify critical domains of basic random variables (the source of randomness) that trigger the patterns. The original response (and its complexity) can be expressed as a probabilistic reconstruction of the identified performance patterns. 

The idea of establishing global characterizations on behaviors of stochastic mechanical systems has been also investigated in the past. For example, in the study of stochastic differential equations, concepts as random attractors and invariant manifolds are developed as global characterizations on stochastic systems \cite{Crauel1997Random}\cite{Roberts2006The}. However, definitions and identifications of random attractors or invariant manifolds involve sophisticated (and often delicate) mathematical considerations, and applications of these concepts to real engineering problems are rare, and generally difficult to cast in a non-intrusive framework. Moreover, for stochastic systems without random attractors or invariant manifolds, they still may exhibit qualitatively different behaviors subjected to certain domains of random input, and thus there are needs to systematically analyze these behaviors.

Another, yet untypical, example for global characterization of stochastic mechanical systems is the concept of mutually exclusive and collectively exhaustive (MECE) set in system reliability theory \cite{LPBounds}\cite{Song2009System}. In a system reliability approach to stochastic mechanical problems, the state space of a mechanical system is partitioned into various performance state levels (e.g., failure or safe in a 2-level partition), and the performance state of the system is contributed by combinations of performance states of components. The MECE set of a system performance state is a set of component performance states to give rise to the system performance state, and the set is MECE. Simply put, the MECE set of a system performance state corresponds to qualitatively different ways (with respect to definitions of components and their performance states) to achieve a system performance state. The limitation of the MECE set concept is that it is useful only if behaviors of a mechanical system can be meaningfully decomposed as combinations of behaviors of components, and such a decomposition should be a prior knowledge. By contrast, the concept of performance-patterns developed in this study is independent of decomposition of the system. In general, compared with random attractors, invariant manifolds, and MECE set, the concept of performance-pattern is more flexible and has promising potential as an effective analysis framework of a large variety of stochastic (not necessarily mechanical) systems. 

In the context of structural reliability, there is another original contribution, which attempts to extract physical characterizations from the performance of a stochastic mechanical system. Starting from the idea of critical excitation \cite{koo2005design}, Fujimura and Der Kiureghian \cite{fujimura2007tail} developed the Tail Equivalent Linearization Method (TELM) to study the reliability of hysteretic mechanical systems. The method, however, goes beyond the statistics of QoI, and it provides a full characterization of the mechanical system in terms of a nonparametric Green's function, the critical excitation (named design point excitation), and the associated design point response. The method later has been proved to be successful in several applications (e.g., \cite{garre2010tail}\cite{broccardo2014further}\cite{alibrandi2017equivalent}). However, TELM is intrusive, and confined to a particular range of systems (e.g., softening and nondegradable systems, and first-order differentiable systems with respect to the input random variables). On the other hand, PPPD is free from these limitations, and generalizes the original idea of TELM for multiple patterns and generic mechanical systems. 

It is essential to remark also that the goal of the proposed framework is fundamentally different from classical sensitivity analysis (i.e., one class of  UQ inverse problems). In sensitivity analysis, the goal is to determine which random variable of the input contributes the most to the variability of the QoIs. Despite being an essential technique to understand the system behavior, it is still incomplete from a physical perspective since it does not highlight the physical patterns underlying the probabilistic structure of the QoIs.

Finally, it did not escape to our attention that the proposed framework can be used in a fully data-driven fashion. In this case, large datasets (of real or synthetic data) of input and output are used to discover patterns and regularities and to build data driven models.

The structure of this paper is as follows. Section \ref{Concepts} introduces the general concepts for the proposed framework. Section \ref{MainTheory} and Section \ref{Computation} respectively develops the theoretical and computational frameworks of probabilistic performance-pattern decomposition (PPPD). Section \ref{Origin} briefly discusses the nature and origins of performance-patterns. Section \ref{NumericalEx} applies the developed methods to the analysis of various mechanical stochastic systems. Finally, Section \ref{Conclusion} presents a series of concluding remarks and future research directions.

\section{General principles of PPPD}\label{Concepts}
\noindent
Consider a mechanical system with a finite\footnote{If a system with an infinite number of random variables is of interest (e.g. systems involve random processes/fields), for practical purposes one could discretize the random processes/fields by a finite set of random variables.} set of basic random variables, denoted by $\bm X$. The basic random variables correspond to the source of randomness\footnote{The definition for ``source of randomness'' is subjected to confinement on the specific physical/mathematical models used to describe the problem.} for the system being considered. In general, the basic random variables involve epistemic uncertainty as well as aleatory variability present within the system or/and external excitation. For classical mechanical systems $\bm X$ may include variables of material properties, geometric quantities, initial and boundary conditions, dynamic excitation, environmental effects, etc. The complete description of stochastic dynamic systems is given by the joint probability distribution of the state variables (i.e., momentum and position) of all degree-of-freedom. However, in engineering applications the system behavior is usually (and better) described by a finite set of response variables, here denoted by $\bm Y$ (which are function of state variables). In this study, response variables are considered instead of state variables because response variables (by definition) are a direct description on the engineering behavior of interest. Since the source of randomness is captured by the basic random variables, the random response variables are deterministic function of basic random variables, i.e.
\begin{equation}
    \bm Y=\mathcal{M}(\bm X)\,,\label{ModelFunc}
\end{equation}
where the model function $\mathcal{M}(\cdot)$ typically stems from fundamental deterministic physical laws (e.g., symmetries/conservation laws). In general, Eq.\eqref{ModelFunc} defines a nonlinear mapping (not necessarily injective) from $\bm X$ to $\bm Y$, and the dimensions of $\bm X$ and $\bm Y$ are generally different. Note that for time variant systems we consider the variable time to be included in definitions of $\bm X$ and $\bm Y$ (e.g., $\bm X$ and $\bm Y$ can represent discretized stochastic processes).

In the context of engineering applications, it is also meaningful to introduce the performance state of the response variables. For example, in the design and safety assessment of civil and mechanical structures, it is vital to know how structures behave under different load and structural conditions (including extreme/rare events). For these cases, a performance state can be introduced to focus on critical domains of the response variables. More abstractly, the performance state of the response variables $\bm Y$ is defined as an event, denoted by $\mathcal{P}_y$, such that $\mathcal{P}_y\subseteq\Omega_y$, where $\Omega_y$ denotes the sample space of $\bm Y$\footnote{Observe that the introduction of $\mathcal{P}_y$ will not influence the generality of this study, one could set $\mathcal{P}_y=\Omega_y$ if the whole sample space is of interest.}. Note that the subscript ``$y$'' in $\mathcal{P}_y$ is introduced to highlight that the performance state is defined in the sample space of $\bm Y$. Then, in the sample space of $\bm X$, denoted by $\Omega_x$, we define the event $\mathcal{P}_x\equiv\left\lbrace \bm{x}|\mathcal{M}(\bm x)\in\mathcal{P}_y,x\in\Omega_x\right\rbrace$.  Specifically, $\mathcal{P}_x$  is the event in the basic random variables space that maps into the event $\mathcal{P}_y$ in the response random variables space. Provided these dual domains $(\mathcal{P}_x, \mathcal{P}_y)$ it is of interest to determine the conditional probability distribution of $\bm Y|\mathcal{P}_y$.

Provided with the joint probability density function (PDF) of the basic random variables $\bm X$, denoted by $f_{\bm{X}}(\bm x)$, the joint PDF of $\bm X$ conditional on $\mathcal{P}_x$ is

\begin{equation}
f_{\bm{X}}(\bm x|\mathcal{P}_x)=\frac{I(\bm x\in\mathcal{P}_x)f_{\bm{X}}(\bm x)}{\int_{\Omega_x}I(\bm x\in\mathcal{P}_x)f_{\bm{X}}(\bm x)\,d\bm x}\,,\label{PDFx}
\end{equation}
where $I(\bm x\in\mathcal{P}_x)$ is a ``hard classifier''  which gives ``1'’ if $\bm x\in\mathcal{P}_x$, and ``0'' the otherwise. Using Eq.\eqref{ModelFunc}, the joint PDF of response variables $\bm Y$ conditional on the performance state $\mathcal{P}_y$ can be expressed by the multiple integral
\begin{equation}
f_{\bm{Y}}(\bm y|\mathcal{P}_y)=\int_{\Omega_x}\delta(\bm y-\mathcal{M}(\bm x))f_{\bm{X}}(\bm x|\mathcal{P}_x)\,d\bm x\,,\label{PDFy}
\end{equation}
where $\delta(\cdot)$ is the Dirac-Delta function. 
%
%

If the joint PDF $f_{\bm{Y}}(\bm y|\mathcal{P}_y)$ could be obtained from Eq.\eqref{PDFy}, a complete statistical description on the response variables $\bm Y$ within a specified performance state is available. However, for nontrivial problems (e.g., problems with $\mathcal{M}(\cdot)$ being nonlinear and computationally demanding, and/or dimensionality of $\bm X$ or $\bm Y$ being high), a direct evaluation of Eq.\eqref{PDFy} is infeasible. As a consequence, for nontrivial problems, instead of attempting to obtain the joint PDF $f_{\bm{Y}}(\bm y|\mathcal{P}_y)$, a common practice is to study statistical properties of $\bm Y$ (given $\bm Y\in\mathcal{P}_y$) using mean, covariance matrix, marginal distributions, and other statistical measures that are relatively convenient to obtain. Depending on the context of application, the statistical measures of interest could vary.

In this study, an alternative path to systematically investigate the probabilistic structure of the response variables $\bm Y$ (within a performance state) is explored. Moreover, this study goes beyond a statistical characterization of the response variables, since the critical domain of the random input (i.e., basic random variables $\bm X$) that generates the probabilistic structure of $\bm Y$ will also be investigated. Specifically, given the performance state of interest, this paper studies the procedure of (a) determining meaningful patterns for response variables, and (b) determining the critical domain of basic random variables that triggers each pattern. This procedure is defined as Probabilistic Performance Pattern Decomposition (PPPD).

\begin{figure}[H]
  \centering
  \includegraphics[scale=0.37]{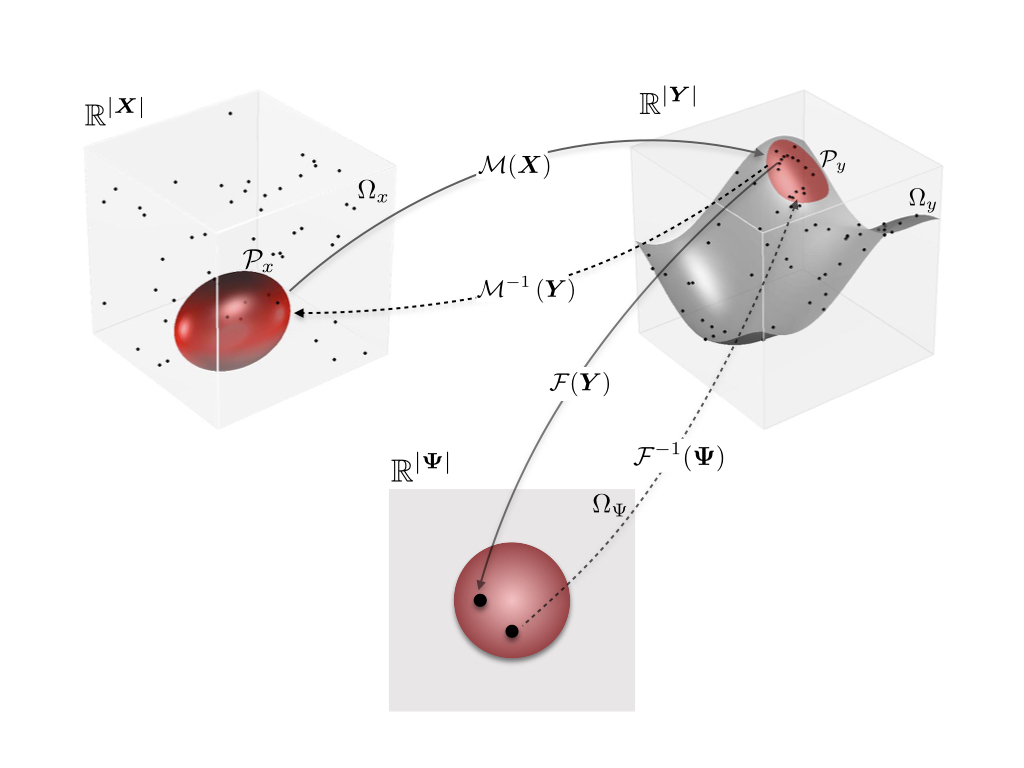}
  \caption{Principles of PPPD. \textit{Direct arrows denotes direct transformations, dashed lines denotes inverse reconstructions}.}
  \label{fig:Intro}
\end{figure}

Figure \ref{fig:Intro} provides a general picture on the PPPD, which particularly shows that PPPD analysis involves the interplay between the basic variables space and the response variables space.
Moreover, we introduce an additional space, namely the feature space, to uncover performance patterns. The basic ingredients of PPPD are described as follows. a) The basic random variables $\bm X$ are mapped to the response variables $\bm Y$ through a model function $\mathcal{M}(\cdot)$. Since a performance state is of particular interest, PPPD focuses on $\mathcal{P}_x$ in the space of $\bm X$ and $\mathcal{P}_y$ in the space of $\bm Y$ ($\mathcal{P}_x$ and $\mathcal{P}_y$ are marked as red regions in the figure). b) The structure of $\mathcal{M}(\bm x|\bm x\in\mathcal{P}_x)$ is typically well-hidden in its embedded (possibly high-dimensional) Euclidean $\bm Y$ space. To disclose its structure, we perform manifold learning via constructing a nonlinear feature projection $\mathcal{F}:\bm Y\longrightarrow\bm\psi$ to identify meaningful patterns. The patterns identified in the feature space are then mapped back to the $\bm Y$ space. c) With the knowledge of the performance patterns in the space of $\bm Y$, we finally identify the critical regions which trigger each pattern in the space of $\bm X$.
\section{Mathematical formulations of probabilistic performance-pattern decomposition (PPPD)}\label{MainTheory}
\subsection{Probabilistic decomposition}
\noindent
We investigate the structure of $f_{\bm{Y}}(\bm y|\mathcal{P}_y)$ by introducing a set of latent random variables\footnote{In physics latent variables are sometimes introduced to reflect the tangible effects of hidden mechanisms which are difficult to observe (but in principle can be observed). In this paper, however, the latent variables are introduced to reflect the abstract concept of functional structure of $f_{\bm{Y}}(\bm y|\mathcal{P}_y)$.} $\bm Z$ defined in an auxiliary sample space $\Omega_z$ with distribution function $Q(\bm z)$. We define $Q(\bm z)$ as the latent distribution and construct a joint distribution between the vector $\bm Y\in\mathcal{P}_y$ and the latent variables $\bm Z$ defined in the augmented sample space $\Omega_{\mathcal{P}_y}\times\Omega_z$ (where we introduce $\Omega_{\mathcal{P}_y}$ to denote the sample space of the performance state $\mathcal{P}_y$). It follows that $f_{\bm{Y}}(\bm y|\mathcal{P}_y)$ can be written as

\begin{equation}
f_{\bm{Y}}(\bm y|\mathcal{P}_y)=\mathbb{E}_{\bm{Z}}\left[f_{\bm{Y}}(\bm y|\bm z;\mathcal{P}_y)\right]=\int_{\bm Z}f_{\bm{Y}}(\bm y|\bm z;\mathcal{P}_y)\,dQ(\bm z)\,,\label{PDFDecom}
\end{equation}
where $\mathbb{E}_{\bm{Z}}[\cdot]$ denotes expectation with respect to the latent variables. Observe that Eq.\eqref{PDFDecom} can be interpreted as the Fredholm integral equation of the first kind, where $f_{\bm{Y}}(\bm y|\bm z;\mathcal{P}_y)$ is the kernel function. Now consider a partition of $\Omega_z$ into a finite set of $K$ mutually exclusive and collective exhaustive events, i.e., $\Omega_z=\cup_{k=1}^K E_k^z$, $E_k^z\cap E_l^z=\varnothing$, $k\neq l$, and $K\in\mathbb{N}^+$, and define $\lambda_k\equiv\mathbb{P}(\bm Z\in E_k^z)$  and $f_{\bm Y}(\bm y|k;\mathcal{P}_y)\equiv f_{\bm Y}(\bm y|\bm z\in E_k^z;\mathcal{P}_y )$. Given this, Eq.\eqref{PDFDecom} can be rewritten to

\begin{equation}
f_{\bm{Y}}(\bm y|\mathcal{P}_y)=\mathbb{E}_{\bm{Z}}\left[f_{\bm{Y}}(\bm y|\bm z;\mathcal{P}_y)\right]=\sum_{k=1}^K \lambda_k f_{\bm{Y}}(\bm y|k;\mathcal{P}_y)\,.\label{PDFDecom1}
\end{equation}
The density $f_{\bm{Y}}(\bm y|k;\mathcal{P}_y)$ is defined as the $k$-th component density and $\lambda_k$, $\sum_{k=1}^K \lambda_k=1$, $\forall k$, $\lambda_k>0$, is defined as the $k$-th component weight. The component density $f_{\bm{Y}}(\bm y|k;\mathcal{P}_y)$ is the likelihood of the realization of $\bm Y$ conditional to the event $\bm Z\in E_k^z$ and performance state $\mathcal{P}_y$, while the component weight $\lambda_k$ provides a direct measure on the importance of $f_{\bm{Y}}(\bm y|k;\mathcal{P}_y)$. Note that $K$ is generally unknown and to be determined in the PPPD procedure. Although Eq.\eqref{PDFDecom} and Eq.\eqref{PDFDecom1} are equivalent, Eq.\eqref{PDFDecom1} offers the advantage of highlighting the decomposition of $f_{\bm Y}(\bm y|\mathcal{P}_y )$ into a finite set of discrete weighted component densities.

It is important to note that different from mixture model approximation to distribution functions, Eq.\eqref{PDFDecom1} is by construction exact. Although the structure of $f_{\bm{Y}}(\bm y|\mathcal{P}_y)$ is an unknown to be disclosed, we assume the structure exists in the space of abstract latent variables. Provided Eq.\eqref{PDFDecom1} to be a formalization for the concept of performance pattern, and given $f_{\bm Y}(\bm y|\mathcal{P}_y )$ to be decomposed in a conceptually meaningful way\footnote{By ``conceptual meaningful'', we indicate that the resulting performance patterns are manifestly different from each other.}, we define the component density $f_{\bm{Y}}(\bm y|k;\mathcal{P}_y)$ to be the PDF of the $k$-th performance pattern, the component weight $\lambda_k$ to be the relative importance of the $k$-th performance pattern, and event $E_k^z$ to be the label of the $k$-th performance pattern. Moreover, the mean of $f_{\bm{Y}}(\bm y|k;\mathcal{P}_y)$ can be regarded as a characteristic vector to represent the performance pattern. Note that although Eq.\eqref{PDFDecom1} provides no hints on how to decompose $f_{\bm Y}(\bm y|\mathcal{P}_y )$ for a specific application, we are interested in the nontrivial cases for which $K>1$ and $f_{\bm Y}(\bm y|k;\mathcal{P}_y )\neq f_{\bm Y}(\bm y|\mathcal{P}_y )$.

For a given realization of $y^{(i)}$ of response variables, we say $y^{(i)}$ belongs to the $k$-th performance pattern with the likelihood

\begin{equation}
\mathcal{L}_k(y^{(i)})=\frac{\lambda_k f_{\bm{Y}}(\bm y^{(i)}|k;\mathcal{P}_y)}{\sum_{j=1}^K\lambda_j f_{\bm{Y}}(\bm y^{(i)}|j;\mathcal{P}_y)}\,.\label{Likelihood}
\end{equation}
The likelihood $\mathcal{L}_k(y^{(i)})$ can be zero if $y^{(i)}\not\in\Omega_{\mathcal{P}_y,k}$, where $\Omega_{\mathcal{P}_y,k}$ denotes the sample space of $f_{\bm{Y}}(\bm y^{(i)}|k;\mathcal{P}_y)$. A zero likelihood also implies $f_{\bm{Y}}(\bm y^{(i)}|k;\mathcal{P}_y)$ is a truncated distribution, i.e. $\Omega_{\mathcal{P}_y,k}\subset\Omega_{\mathcal{P}_y}$. If the component densities are truncated distributions, the performance patterns provide a ``hard decomposition'' (partition) of $\mathcal{P}_y$, otherwise they provide a ``soft decomposition'' in which each realization of $f_{\bm Y}(\bm y|\mathcal{P}_y)$ has a nonzero probability to belong to any of the patterns.

Now, to identify the critical domains of basic random variables that trigger each performance pattern, using a change of variables we obtain

\begin{equation}
f_{\bm X}(\bm x|k;\mathcal{P}_x)=\frac{f_{\bm{Y}}(\mathcal{M}(\bm x)|k;\mathcal{P}_x)}{\int_{\Omega_x}f_{\bm{Y}}(\mathcal{M}(\bm x)|k;\mathcal{P}_x)\,d\bm x}\,,\label{GenerateDensity}
\end{equation}
where $f_{\bm X}(\bm x|k;\mathcal{P}_x)$ is named the generating density for the $k$-th performance pattern. Similar to Eq.\eqref{PDFDecom1}, $f_{\bm X}(\bm x|\mathcal{P}_x)$ can be written in the decomposition form

\begin{equation}
f_{\bm{X}}(\bm x|\mathcal{P}_x)=\sum_{k=1}^K \lambda_k f_{\bm{X}}(\bm x|k;\mathcal{P}_x)\,.\label{GDensityDecomp}
\end{equation}
Note that the $\lambda_k$ in Eq.\eqref{PDFDecom1} and Eq.\eqref{GDensityDecomp} are, by definition, identical.

\subsection{Feature space representation}
\noindent Eq.\eqref{PDFDecom1} should be constructed such that the performance patterns are ``manifestly different'' from each other. To define manifestly different, we first introduce the feature mapping of $\bm Y$ described as

\begin{equation}\label{FeatureMap}
\begin{array}{lr}
\bm\Psi=\mathcal{\bm F}(\bm Y) \\
\hat{\bm Y}=\mathcal{\bm F}^{-1}(\bm\Psi)
\end{array}
\end{equation}
where the dimensionality of the feature vector $\bm\Psi$ is typical much smaller than $\bm Y$. Note that $\mathcal{\bm F}^{-1}$ represents the reconstruction function rather than the inverse function, and typically the inverse function does not exist since the feature mapping is in general not bijective\footnote{In feature mapping, typically, there is a compression and loss of information, therefore the reconstruction is partial.}. The feature mapping is introduced to disclose the structure of $f_{\bm{Y}}(\bm y|\mathcal{P}_y)$, and in the feature space similar to Eq.\eqref{PDFDecom1} the projected decomposition is

\begin{equation}
f_{\bm{\Psi}}(\bm\psi|\mathcal{P}_{\psi})=\sum_{k=1}^K\lambda_k f_{\bm{\Psi}}(\bm\psi|k;\mathcal{P}_{\psi})\,.\label{PDFDecomFeat}
\end{equation}

A natural requirement for the projected performance patterns $f_{\bm{\Psi}}(\bm\psi|k;\mathcal{P}_{\psi})$ is: the expected within-pattern distance should be smaller than the expected between-pattern distance, i.e.
\begin{equation}
\int d(\bm\psi,\bm\psi')f_{\bm{\Psi}}(\bm\psi|k;\mathcal{P}_{\psi})f_{\bm{\Psi}}(\bm\psi'|k;\mathcal{P}_{\psi})\,d\psi d\psi'<\int d(\bm\psi,\bm\psi'')f_{\bm{\Psi}}(\bm\psi|k;\mathcal{P}_{\psi})f_{\bm{\Psi}}(\bm\psi''|l;\mathcal{P}_{\psi})\,d\psi d\psi''\,,\label{PatternDistance}
\end{equation}
where $k\not=l$, and $d(\cdot)$ is a specified distance measure. Eq.\eqref{PatternDistance} simply states that the within-pattern similarity should be larger than the between-pattern similarity, and it provides a guidance on constructing performance patterns. One should note that Eq.\eqref{PatternDistance} does not address mathematical issues such as well-posedness (existence, uniqueness, and stability), which are outside the scope of the current study.

\section{Computational framework of PPPD}\label{Computation}

\subsection{Realizations of basic and response random variables}\label{RandomRealization}
\noindent
 In this section, we introduce the computational framework for PPPD based on sampling methods. Specifically, the framework is developed using a dataset of random realizations of $\bm X$ drawn from PDF $f_{\bm X} (\bm x|\mathcal{P}_x)$, and the ``corresponding'' realizations of $\bm Y$ (by ``corresponding'', we indicate that Eq.\eqref{ModelFunc} is satisfied for each pair of $\bm X$ and $\bm Y$). Note that for random samples of $\bm X$ drawn from PDF $f_{\bm X} (\bm x|\mathcal{P}_x)$, the corresponding samples of $\bm Y$ naturally follow $f_{\bm Y} (\bm y|\mathcal{P}_y)$. 

Before developing methods to sample from $f_{\bm X} (\bm x|\mathcal{P}_x)$, it is useful to introduce notations of the limit-state surface to describe boundary of the performance state $\mathcal{P}_x$. Let the limit-state surface be written as

\begin{equation}
G(\bm y)=0\,.\label{LimitStateY}
\end{equation}
Moreover, $G(\bm y)\leq0$ denotes $\bm y$ within the performance state $\mathcal{P}_y$, and $G(\bm y)>0$ denotes the otherwise; then, $\mathcal{P}_y$ can be written as

\begin{equation}
\mathcal{P}_y=\left\lbrace\bm y|G(\bm y)\leq0\right\rbrace\,.\label{Py}
\end{equation}
Using Eq.\eqref{ModelFunc} $\mathcal{P}_x$ can be written as

\begin{equation}
\mathcal{P}_x=\left\lbrace\bm x|G(\mathcal{M}(\bm x))\leq0\right\rbrace\,.\label{Px}
\end{equation}

A na\"ive rejection sampling based approach could be applied to generate random samples from $f_{\bm X}(\bm x|\mathcal{P}_x)$ such that it continues drawing samples from $f_{\bm X}(\bm x)$ and only saves the ones with $G(\mathcal{M}(\bm x))\leq0$. The na\"ive rejection sampling approach is effective if $\mathbb{P}(\bm X\in\mathcal{P}_x)$ (or $\mathbb{P}(\bm Y\in\mathcal{P}_y)$ equivalently) is relatively large. However, if $\bm X\in\mathcal{P}_x$ is characterized as a rare event, a large majority of samples would fall outside the performance state, consequently the na\"ive rejection sampling approach becomes practically infeasible.

For rare event simulations, one attractive approach with wide applicability is the sequential Monte Carlo (SMC) /Subset Simulation method \cite{cerou2012sequential}\cite{moral2006sequential}\cite{Au:2001aa}. A key concept in the SMC approach to sample from $f_{\bm X}(\bm x|\mathcal{P}_x)$ is to introduce a finite sequence of intermediate states, denoted by $\mathcal{P}_x^{(j)}$, $j=1,2,...,m$, that satisfies

\begin{equation}
\mathcal{P}_x^{(1)}\supset\mathcal{P}_x^{(2)}\supset\cdot\cdot\cdot\supset\mathcal{P}_x^{(m)}=\mathcal{P}_x\,.\label{PSequence}
\end{equation}

One approach to construct $\mathcal{P}_x^{(j)}$ that satisfies Eq.\eqref{PSequence} is to introduce a sequence of parameters $g^{(j)}$ such that $\mathcal{P}_x^{(j)}$ is expressed by

\begin{equation}
\mathcal{P}_x^{(j)}=\left\lbrace\bm x|G(\mathcal{M}(\bm x))-g^{(j)}\leq0\right\rbrace\,,\label{Pj}
\end{equation}
where $g^{(j)}$ is monotonic with $j$, i.e. $g^{(1)}>g^{(2)}>\cdot\cdot\cdot>g^{(m)}=0$.

The intermediate states in SMC can be either prespecified using certain rule of thumbs \cite{moral2006sequential} or selected adaptively such that the probability $\mathbb{P}(\bm X\in\mathcal{P}_x^{(j+1)}|\bm X\in\mathcal{P}_x^{(j)})$ lies in a proper range \cite{cerou2012sequential}\cite{Au:2001aa}. In general, compared with a fixed intermediate states approach, an adaptively selected sequence of intermediate states would lead to more effective SMC sampling. In each intermediate step of the SMC simulation, a Markov Chain Monte Carlo (MCMC) sampler is performed starting with seed samples within $\mathcal{P}_x^{(j)}$ to generate samples for $\mathcal{P}_x^{(j+1)}$. An SMC procedure that adaptively specify $\mathcal{P}_x^{(j)}$ to sample from $f_{\bm X}(\bm x|\mathcal{P}_x)$ is described in \ref{AppendSampling}.

\subsection{Feature mapping via manifold learning}
\noindent
By applying the aforementioned Monte Carlo methods, one would obtain a dataset consisting of $N$ pairs of $\bm X$ and $\bm Y$ samples that follow $f_{\bm X}(\bm x|\mathcal{P}_x)$ and $f_{\bm Y}(\bm y|\mathcal{P}_y)$, respectively. Let $\mathcal{\bm Y}=[\bm y^{(1)},…,\bm y^{(N)}]$ denote the dataset of $N$ samples of $\bm Y$. Ideally, one should be able to identify patterns in $\mathcal{Y}$. However, analysis directly on $\mathcal{Y}$ could encounter significant challenges if the dimensionality of $\bm Y$ is high and/or topologies of $\mathcal{P}_y$ is complex (e.g., $\bm Y$ is associated with random processes/fields generated from some complex physics mechanism). As introduced in Section \ref{MainTheory}, we apply feature mapping Eq.\eqref{FeatureMap} to cast samples of $\bm Y$ into a low dimensional feature space. Note that if $\mathcal{\bm Y}$ involves components from different sources with different scales, it can be beneficial to perform normalization before the feature space transformation.

In this paper, two nonlinear feature mappings based on manifold learning will be investigated.

\subsubsection{Diffusion map}
\noindent
Diffusion map \cite{coifman2005geometric}\cite{Coifman2006Diffusion} is a manifold leaning method that uses eigen-functions of a Markov matrix (describing affinities in a dataset) to generate informative and simplified representations of a dataset. As a manifold learning technique the diffusion map can be used to discover the underlying manifold that the data has been sampled from. Moreover, in diffusion map the diffusion distance defined as the Euclidean distance in the embedding space is a robust and noise-insensitive metric reflecting the connectivity of the dataset \cite{coifman2005geometric}. Implementation details of the diffusion map for PPPD can be found in \ref{AppendDMap}.

\subsubsection{Autoencoder}
\begin{figure}[H]
  \centering
  \includegraphics[scale=0.32]{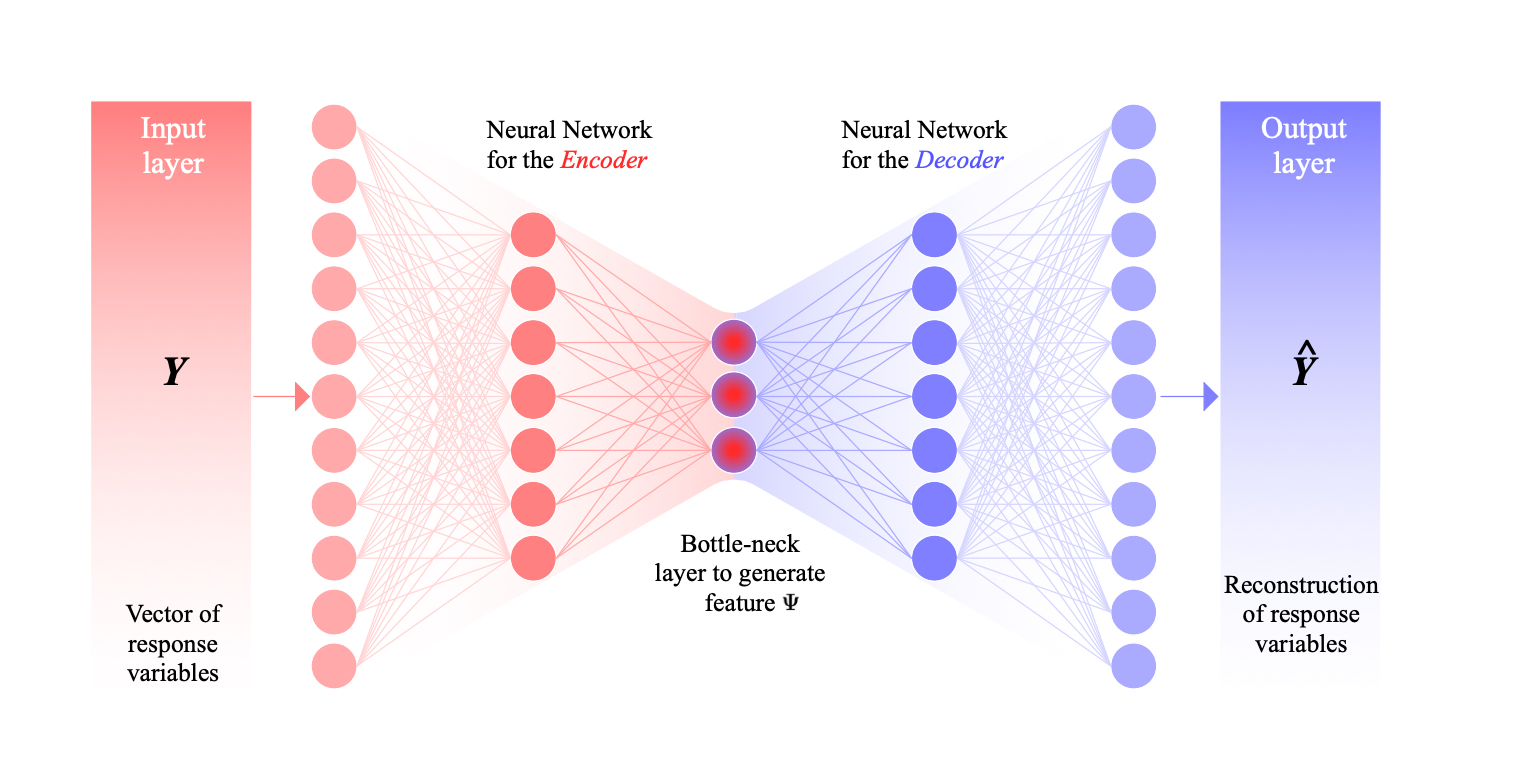}
  \caption{Autoencoder in PPPD analysis. \textit{The basic idea of autoencoder is: the output of bottleneck layer must contain main structure of the original input, otherwise the reconstruction cannot be satisfactory.}}
  \label{fig:Auto}
\end{figure}

\noindent
The autoencoder \cite{hinton2006reducing}\cite{vincent2010stacked} is a manifold leaning method that uses feed-forward neural network to generate simplified encoding of a dataset. In the context of PPPD analysis, an autoencoder consists of an encoder which maps each response vector $\bm y^{(i)}$ into a feature vector $\bm\psi^{(i)}$, and a decoder which maps $\bm\psi^{(i)}$ back to a reconstruction of $\bm y^{(i)}$, denoted as $\hat{\bm y}^{(i)}$. The autoencoder is trained to minimize the distance between $\bm y^{(i)}$ and $\hat{\bm y}^{(i)}$, i.e. the reconstruction error. The basic concept of the autoencoder in PPPD analysis is illustrated in Figure \ref{fig:Auto}. 

Owing to the flexibility of artificial neural network techniques, compared with the diffusion map autoencoder can be more attractive in processing complex dataset. Implementation details of the autoencoder for PPPD can be found in \ref{AppendAutoEncoder}.
\subsection{Performance pattern identification}\label{PatRec}
\noindent
Given the set of feature vectors $\bm\Psi$, the subsequent step of PPPD is to find patterns in $\bm\Psi$. A Monte Carlo discretization of Eq.\eqref{PatternDistance} naturally leads to the following problem: find an appropriate grouping of a dataset such that the within-group similarity is larger than the between-group similarity. Provided that the correct structure of $f_{\bm Y} (\bm y|\mathcal{P}_y)$ is described by the augmented space of latent variables $\Omega_{\mathcal{P}_y}\times\Omega_z$, the problem can be alternatively interpreted as to restore the complete description $(\bm z,\bm Y)$ from samples of $\bm Y$\footnote{Note that $\bm z$ is an abstract vector to represent the structure of $f_{\bm Y} (\bm y|\mathcal{P}_y)$, therefore the numerical value of $\bm z$ is meaningless. Rigorously speaking, it is the indicator $\bm z\in E_k^z$ to be restored.}. This is a well-known unsupervised statistical learning problem that can be tackled by clustering analysis.

\subsubsection{Determine the number of patterns \& Clustering analysis}
\noindent
Normally if the application of manifold learning could effectively map the original samples into a two- or three- dimensional feature space, the number of patterns is expected to be trivially identified. For a relatively high dimensional feature space embedding, to determine the number of patterns one can study a strict partitioning clustering problem described as follows.

Given a dataset $\bm\Psi$ of $N$ samples, find a partition, denoted by $P=\left\lbrace P_1,…,P_{K^*}\right\rbrace$, of the $N$ samples into $K^*$, $K^*\leq N$, subsets so as to minimize a specified measure of the partition.

\begin{equation}
P^*=\mathop{\arg\min}_Pq(P)\,,\label{Clust}
\end{equation}
where the measure $q(\cdot)$ is defined to be independent of $K^*$ so that $K^*$ is also an unknown to be determined from Eq.\eqref{Clust}.

In clustering analysis practice, a two-step approach is typically used to solve Eq.\eqref{Clust}. In the first step, a measure $q_K(P)$ is defined to find the optimal partition for a specified number of clusters. For example, in the classical k-means clustering method $q_K(P)$ is defined by the within-cluster sum of squares, i.e.

\begin{equation}
q_K(P)=\sum_{i=1}^{K}\sum_{\bm\psi^{(j)}\in P_i}\left\|\bm\psi^{(j)}-\bm\mu_i\right\|^2\,,\label{qK}
\end{equation}
where $\bm\mu_i$ is the mean of $\psi^{(j)}$ in $P_i$. With $q_K(P)$ specified, the optimal partition for a specified $K$, denoted as $P_K^*$, is obtained from

\begin{equation}
P_K^*=\mathop{\arg\min}_Pq_K(P)\,.\label{ClustK}
\end{equation}
Even though the optimization problem defined by Eq.\eqref{ClustK} is usually NP-hard, various clustering algorithms \cite{Cluster} have been developed to search for the approximate solutions and proven to be effective for practical applications.

In the second step, a measure $\ell(P_K^*)$ is defined to find the optimal number of clusters, $K^*$, and consequently the optimal partition $P^*$ via

\begin{equation}
\begin{aligned}\label{Kstar}
  &K^*=\mathop{\arg\min}_{K\in\mathbb{N}^+}\left\lbrace \ell(P_K^*)\right\rbrace\\
  &P^*=P_{K=K^*}^*
\end{aligned}
\end{equation}

The specification of $\ell(\cdot)$ belongs to the problem of determining the ``exact'' number of groups in a dataset, which is a fundamental, yet largely unsolved challenge in clustering analysis. Numerous approaches to this problem have been suggested over the past decades \cite{rousseeuw1987silhouettes:}\cite{sugar2003finding}\cite{Amorim2015Recovering}. One attractive approach is based on information theory \cite{sugar2003finding}. In the information theoretic approach, $\ell(\cdot)$ is defined as

\begin{equation}
\ell(P_K^*)=d_{K-1}^{-a}-d_{K}^{-a}\,,\label{InformTheo}
\end{equation}
where the transformation power $a$ is typically set to $a=n/2$, in which $n$ is the dimension, $d_0$ is defined to be 0, and $d_K$ is the approximate distortion expressed by

\begin{equation}
d_{K}=\frac{1}{n\cdot N}\min_{k=1,..,K}\sum_{i=1}^{N}(\bm\psi^{(i)}-\bm\mu_k^{(i)})^T\bm\Sigma^{-1}_k(\bm\psi^{(i)}-\bm\mu_k^{(i)})\,,\label{Distort}
\end{equation}
where $\bm\Sigma_k$ denotes the covariance matrix and $\bm\mu_k^{(i)}$ denotes the cluster center that is closest to sample $\bm\psi^{(i)}$, for a specified $k$.

The essential idea of the information theoretic approach is to use the $K$ versus $\ell(P_K^*)$ curve to investigate the influence of number of clusters on the clustering quality. The distortion $d_K$ is a measure of the within-cluster dispersion, and it is monotonically decreasing as $K$ increases. The information theoretic approach assumes that if $K$ is approaching the ``true'' number of clusters, $K^*$, the drop in distortion will attain the maximum ($\ell(P_K^*)$ will attain the minimum), since past $K^*$ adding more clusters simply partitions within rather than between groups.

Once the partition $P^*$ is obtained, a participation factor of each cluster can be evaluated by

\begin{equation}
\Gamma_k=\frac{\sum_{j=1}^{N}I(\bm\psi^{(j)}\in P_k)}{N}\,,\label{PartF}
\end{equation}
where the indicator function $I(\bm\psi^{(j)}\in P_k)=1$ if $\bm\psi^{(j)}\in P_k$ and $I(\bm\psi^{(j)}\in P_k)=0$ the otherwise. The participation factor can be used as an approximate to the component weight, $\lambda_k$, of each performance pattern. Moreover, the mean vector of each cluster, or the sample closest to the mean vector, can be used as a characteristic vector to represent each performance pattern.

Other than hard clustering approaches, one could also use soft clustering algorithms \cite{Cluster} to establish a soft decomposition. Recalling concepts introduced in Section \ref{MainTheory}, a hard clustering corresponds to a hard decomposition (partition) of $f_{\bm Y}(\bm y|\mathcal{P}_y)$, and each sample can only belong to one of the patterns; while a soft clustering corresponds to a soft decomposition of $f_{\bm Y}(\bm y|\mathcal{P}_y)$, and each sample is allowed to belong to more than one pattern. 	

\subsubsection{Parametric description of performance patterns}
\noindent
Given the results of a clustering analysis, one could construct a parametric model to describe the component densities $f_{\bm Y}(\bm y|k;\mathcal{P}_y )$ for each cluster/performance pattern.

A typical approach to construct a parametric PDF model is to use mixture distribution. Specifically, $f_{\bm Y}(\bm y|\mathcal{P}_y)$ can be written in terms of a parametric mixture model, and $f_{\bm Y}(\bm y|k;\mathcal{P}_y)$ is described by component of the parametric mixture model, i.e.

\begin{equation}
\begin{aligned}\label{MixtureMod}
  &f_{\bm Y}(\bm y|\mathcal{P}_y)\approxeq\sum_{k=1}^{K^*}\hat{\lambda}_k f_{\bm Y}(\bm y|\bm\theta,\bm\theta_k;\mathcal{P}_y)\\
  &f_{\bm Y}(\bm y|k;\mathcal{P}_y)\approxeq f_{\bm Y}(\bm y|\bm\theta,\bm\theta_k;\mathcal{P}_y)
\end{aligned}
\end{equation}
where $\bm\theta$ is a set of global parameters, $\bm\theta_k$ is a set of component parameters, and $\hat{\lambda}_k$ is the component weight of the mixture model. Parameters of the mixture model can be estimated by the Expectation-Maximization (EM) algorithm \cite{dempster1977maximum}, guided by the partition and labeled samples obtained from clustering analysis. Note that the component density $f_{\bm Y}(\bm y|\bm\theta,\bm\theta_k;\mathcal{P}_y)$ in Eq.\eqref{MixtureMod} could also be represented by a mixture model.

Given the component density $f_{\bm Y}(\bm y|\bm\theta,\bm\theta_k;\mathcal{P}_y)$, the generating density $f_{\bm X}(\bm x|k;\mathcal{P}_x)$ can be obtained by Eq.\eqref{GenerateDensity}, in principle. However, since in general the model function $\mathcal{M}(\cdot)$ is not explicit, Eq.\eqref{GenerateDensity} is particularly useful only when a Monte Carlo approach is employed to sample from $f_{\bm X}(\bm x|k;\mathcal{P}_x)$. If a parametric description of $f_{\bm X}(\bm x|k;\mathcal{P}_x)$ is of interest, one could employ the mixture model approach.

Finally, it is important to note that parametric descriptions of $f_{\bm Y}(\bm y|k;\mathcal{P}_y)$ or $f_{\bm X}(\bm x|k;\mathcal{P}_x)$ are not always feasible. For generic problems incapable of parametrization, the numerical solutions obtained from clustering analysis can be regarded as the final output of PPPD analysis. In clustering analysis, instead of a parametric description one could only obtain statistical/geometrical descriptions on each performance pattern and its generating density.

\subsection{Procedures of PPPD}\label{ProcedurePPPD}
\noindent
To conclude the ideas introduced in this section, the basic computational procedures of PPPD analysis is described as follows.

\begin{algorithm}[H]
\caption{Procedures of PPPD analysis}\label{alg:PPPD}
\begin{description}

\item [Step 1: Problem statement]
\rule{0pt}{15pt}
\begin{itemize}
\item Define basic random variables $\bm X$, and define the joint PDF of $\bm X$.
\item Define the response random variables $\bm Y$ to describe the behavior of the system.
\item Specify the computational model $\mathcal{M}(\cdot)$ that maps $\bm X$ to $\bm Y$.
\item Define the performance state of interest.
\end{itemize}

\item [Step 2: Obtain random realizations of basic and response variables]
\rule{0pt}{15pt}
\begin{itemize}
\item Draw $N$ pair of samples from PDFs $f_{\bm X}(\bm x|\mathcal{P}_x)$ and $f_{\bm Y}(\bm y|\mathcal{P}_y)$.
\end{itemize}

\item [Step 3: Feature mapping]
\rule{0pt}{15pt}
\begin{itemize}
\item Perform feature mapping on samples of $\bm Y$.
\end{itemize}

\item [Step 4: Performance pattern identification]
\rule{0pt}{15pt}
\begin{itemize}
\item Determine the number of performance patterns in the feature space.
\item Extract the performance patterns of $\bm Y$ in the feature space and their generating densities $f_{\bm X}(\bm x|k;\mathcal{P}_x)$ via clustering analysis. 
\item (Optional) Obtain a parametric description on performance patterns of $\bm Y$ and their generating densities $f_{\bm X}(\bm x|k;\mathcal{P}_x)$.
\end{itemize}
\end{description}
\end{algorithm}

\section{Origin of performance patterns}\label{Origin}
\noindent
In this Section, we investigate the origin of performance patterns, i.e. the possible causes that generate multiple performance patterns. Clearly, a necessary but not sufficient condition for observing multiple performance patterns is the random variability within the system or/and the external excitation, otherwise the performance of the system will be an individual and deterministic event.

Given that there are randomness involved, the origin of multiple performance patterns can be traced back to the following causes.

\begin{tcolorbox}[colframe=black!5!white]
\begin{center}
\begin{tabular}{l}
a) \textit{Source}: The existence of multiple patterns in the basic random variables.\\
b) \textit{Propagation}: The existence of bifurcations or discontinuities within the deterministic physical\\ 
\ \ \ \ \ \ \ \ \ \ \ \ \ \ \ \ \ \ \ \ \ \  model.\\
c) \textit{Constraint}: The specific property of the performance state.\\
d) \textit{Subjectivity}: The specific property of the distance metric defined in feature mapping.
\end{tabular}
\end{center}
\end{tcolorbox}

To understand ``c) \textit{Constraint}'', note that the performance state $\mathcal{P}_y$ applies a truncation to the original sample space of response variables, and after the truncation the conditional distribution $f_{\bm Y}(\bm y|\mathcal{P}_y)$ could exhibit multiple patterns even if $f_{\bm Y}(\bm y)$ is unimodal. To understand ``d) \textit{Subjectivity}'', note that a redefinition of the distance metric alters the structure of the dataset, so that patterns that are not inherent in the original dataset could be triggered. In the feature mapping procedure, if a conventional distance metric is used (e.g., the Euclidean distance), the manifold learning technique could, at best, make patterns that are ambiguous in the original space easier to be identified. However, if a problem specific distance metric is used, the distance metric introduces additional prior knowledge (subjectivity) so that new patterns (that do not exist within the original dataset) could be triggered. It follows that the use of an inappropriate problem specific distance metric could produce artificial performance patterns which lack conceptual importance, thus the use of problem specific distance metrics should be handled with cautiousness. However, on the other hand, using meaningful physics-informed distance metric may assist the discovery of important well-hidden structures. Investigations on the use of physics-informed distance metric will be addressed in the follow-up studies.

It can be concluded from this section that the performance pattern not only reflects characteristics of the randomness source and deterministic physical model, but also is able to encompass properties of the specific domain of interest and the problem specific understandings on system behaviors. Therefore, the performance pattern can be regarded as a holistic characterization of the stochastic system being studied.

\section{Numerical investigations}\label{NumericalEx}
\subsection{An illustrative example of simple system identification}
\noindent
To illustrate main ideas and procedures of PPPD analysis, consider a hypothetical system with basic random variables $\bm X$ of the form $\bm X=[\bm X_p,X_{st}]$, where $\bm X_p=[X(t_1),…,X(t_n)]$, $X(t_i)\sim \mathcal{N}(0,1)$, $i=1,..,n$, represents a discretized zero-mean Gaussian white noise, and $X_{st}\in\left\lbrace1,2,3,4\right\rbrace$, is a discrete uniform random variable. For a realization of $X_{st}$, the response of the hypothetical system, $\bm Y=[Y(t_1),…,Y(t_n)]$, is a discretized stochastic process expressed by

\begin{equation}
\bm Y=\left\lbrace\begin{aligned}
  &\sin(\pi\bm t/4)+\cos(\pi\bm t/3)+0.3\bm X_p\,,\ X_{st}=1\\
  &\sin(\pi\bm t/4+0.1)+\cos(\pi\bm t/3-0.1)+0.3\bm X_p\,,\ X_{st}=2\\
  &\sin(\pi\bm t/3)+\cos(2\pi\bm t/5)+0.3\bm X_p\,,\ X_{st}=3\\
  &\sin(\pi\bm t/3+0.1)+\cos(2\pi\bm t/5-0.1)+0.3\bm X_p\,,\ X_{st}=4\\
\end{aligned}\right.\label{ModFunEx1}
\end{equation}
where $\bm t=[t_1,..,t_n]$. It is assumed the whole sample space of $\bm Y$ is of interest, i.e. $\mathcal{P}_y=\Omega_y$. The time sequence $\bm t$ is set to starting from 0.01 seconds to 10 seconds, with a uniform incremental time step of 0.01 seconds. Therefore, the dimension of $\bm Y$ is 1000.

\begin{figure}[H]
  \centering
  \includegraphics[scale=0.70]{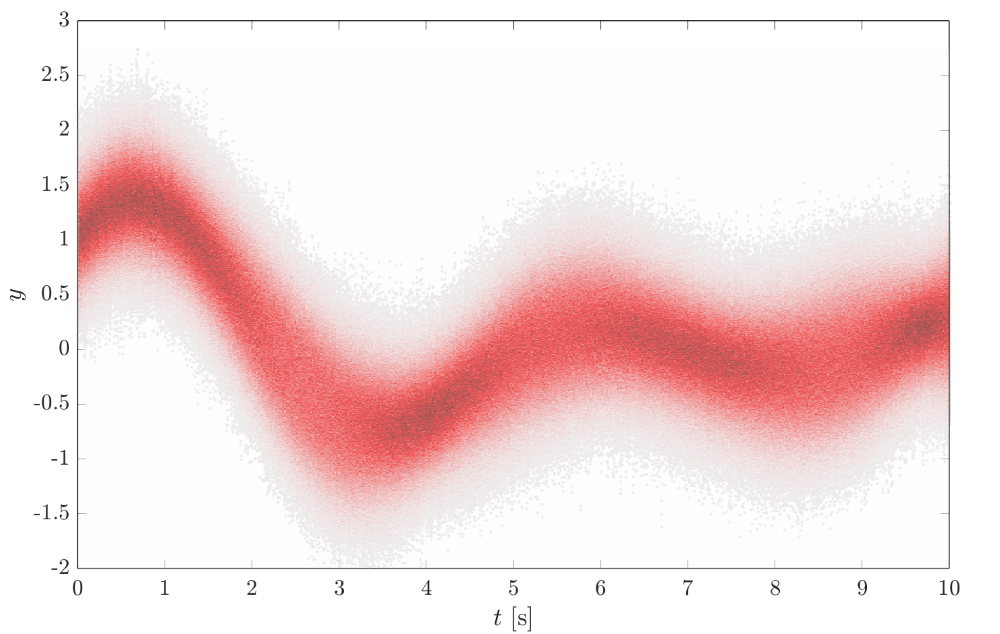}
  \caption{Realizations of response variables in the original space. \textit{The color-map represent the density of the sample points}.}
  \label{fig:Ex1Fig1}
\end{figure}

\begin{figure}[H]
  \centering
  \includegraphics[scale=1.2]{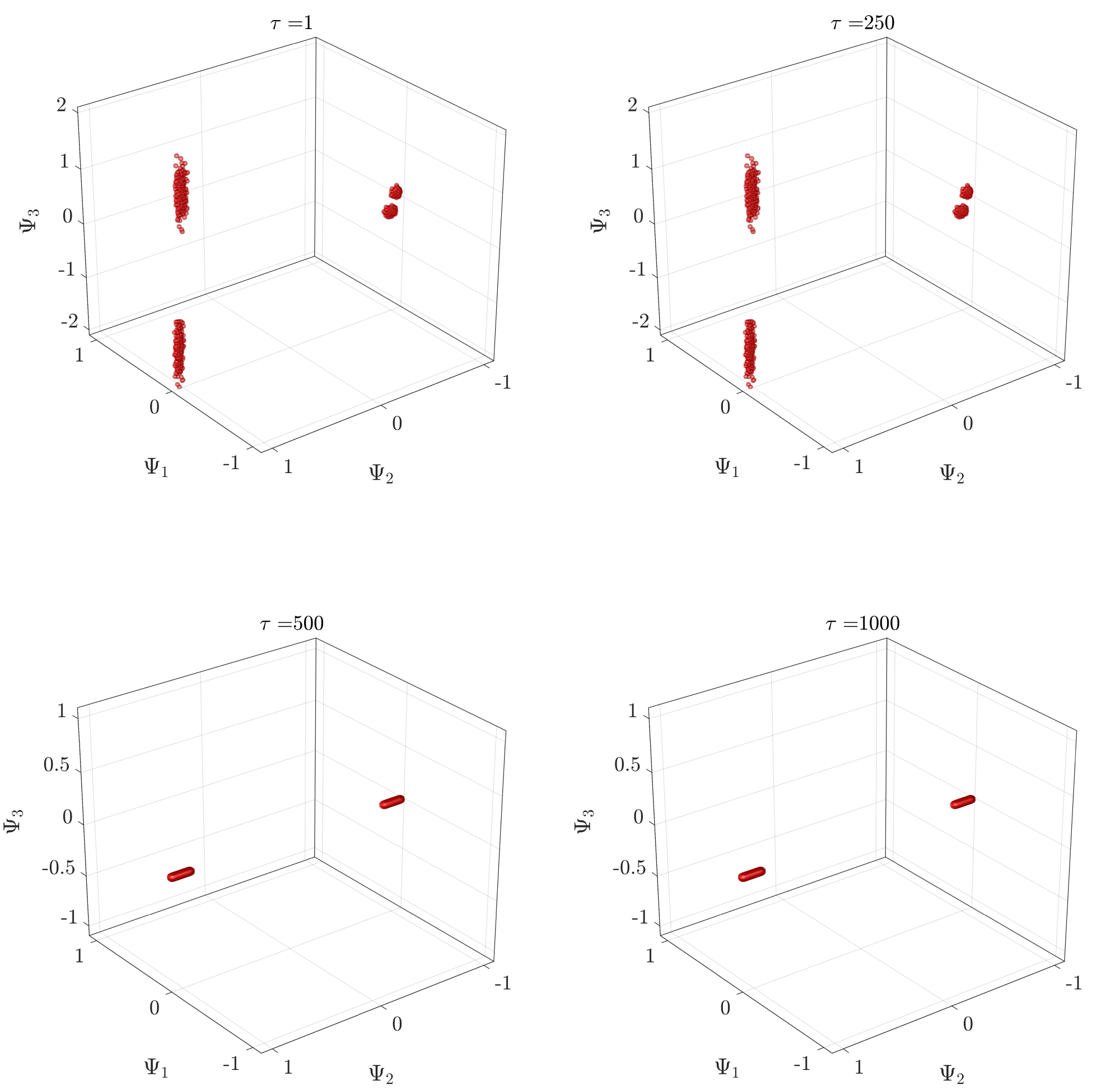}
  \caption{Feature space representation using the diffusion map with various time-scales}
  \label{fig:Ex1Fig2}
\end{figure}

\begin{figure}[H]
  \centering
  \includegraphics[scale=1.2]{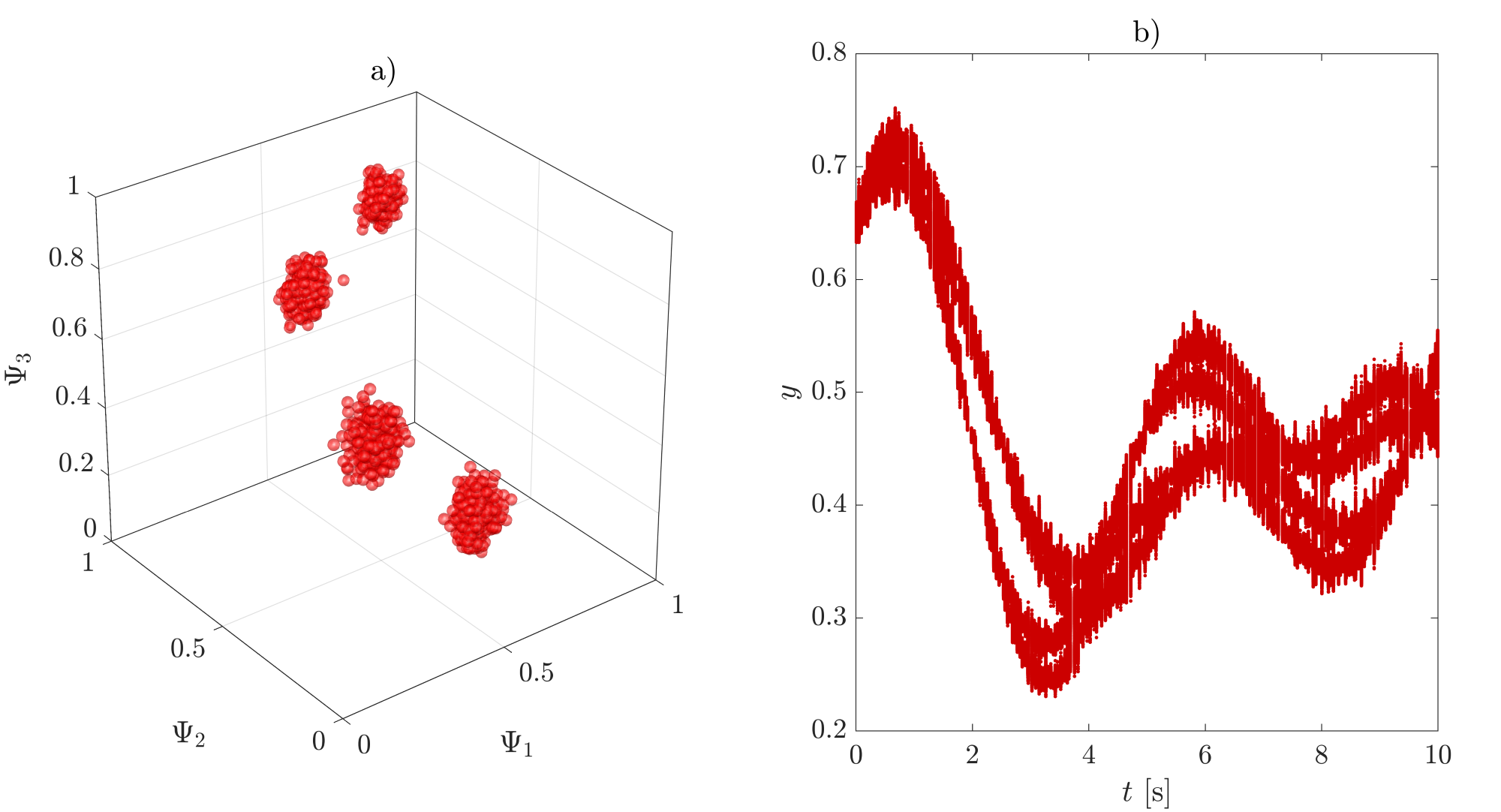}
  \caption{Feature space representation a) and reconstruction b) using the autoencoder}
  \label{fig:Ex1Fig3}
\end{figure}

Now it is assumed one can only observe the input $\bm X$ and output $\bm Y$, without a prior knowledge on Eq.\eqref{ModFunEx1}. The PPPD analysis is used to retrieve structuralized information from the dataset of $\bm X$ and $\bm Y$. To start the PPPD analysis, using a direct Monte Carlo simulation 2000 random realizations of $\bm Y$ are obtained (shown in Figure \ref{fig:Ex1Fig1}). By a visual inspection on Figure \ref{fig:Ex1Fig1} it seems impossible to identify if there is more than one performance pattern.
Figure \ref{fig:Ex1Fig2} shows the 2000 realizations of Y embedded into a 3-dimensional feature space, obtained from diffusion maps with various time-scales $\tau$. The similarity matrix is constructed using Eq.\eqref{SimilarityM} with $\bm L^2$-norm distance and $\epsilon$ is setting to 10. It can be observed from Figure \ref{fig:Ex1Fig2} that: (a) for a relatively high resolution embedding (a relatively small time-scale $\tau$), four patterns can be identified in the feature space; (b) for a relatively low resolution embedding (a relatively large time-scale $\tau$), two patterns can be identified in the feature space. (Note that if $\tau$ is set to be large enough eventually there will be only one pattern, yet this is a trivial case.) This observation implies: (a) there exists four patterns; (b) the four patterns can be divided into two groups, and for each group the similarity within the group is more significant than the similarity between groups.	
Next, an autoencoder with 5 hidden layers and 100-30-3 neurons for each hidden layer of the encoder (the decoder is symmetric) is employed for feature mapping. The sigmoid transfer function is employed for all neurons. The neural network is trained using the scaled conjugate gradient algorithm \cite{moller1993original}, with a mean square error cost function (without sparsity or other regularization terms). Note that prior to training, a min-max normalization is applied to $\bm Y$ (since the output of a sigmoid function lies in $[0,1]$). Figure \ref{fig:Ex1Fig3} shows the 3-dimensional feature space representation and the reconstructed $\bm Y$. Note that to obtain the reconstruction an inverse of the min-max normalization is applied to the output layer.

\begin{figure}[H]
  \centering
  \includegraphics[scale=1.7]{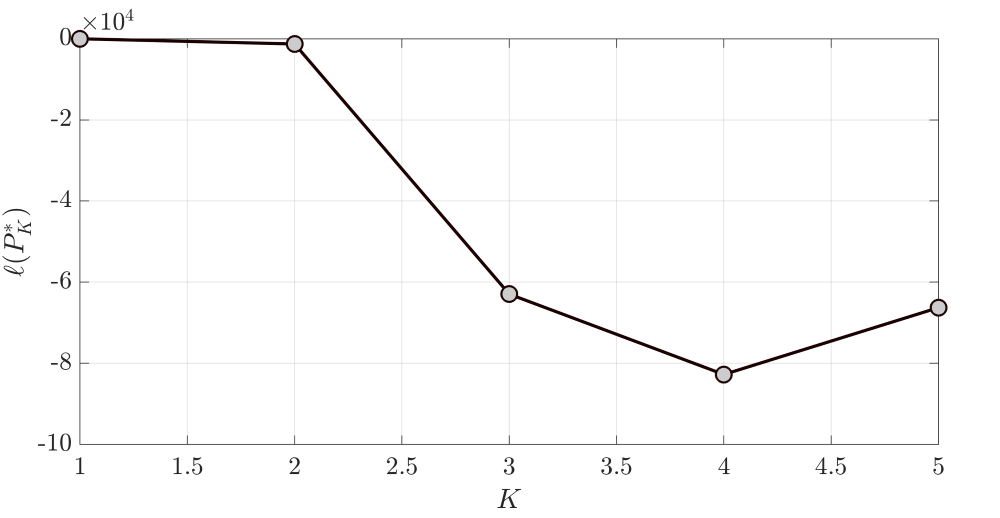}
  \caption{The $\ell(P_K^*)$-$K$ curve of the information theoretic approach}
  \label{fig:Ex1Fig4}
\end{figure}

Although for this example one could visually identify the number of performance patterns, for illustrative purpose, the information theoretic approach is applied to the diffusion map of $\tau=1$. Figure \ref{fig:Ex1Fig4} shows the $\ell(P_K^*)$-$K$ curve obtained from the information theoretic approach. It is seen from the figure that there is an abrupt drop in $K=4$, indicating a significant decrease in the distortion from grouping into three patterns to grouping into four patterns, thus suggesting $K^*=4$.

\begin{figure}[H]
  \centering
  \includegraphics[scale=.485]{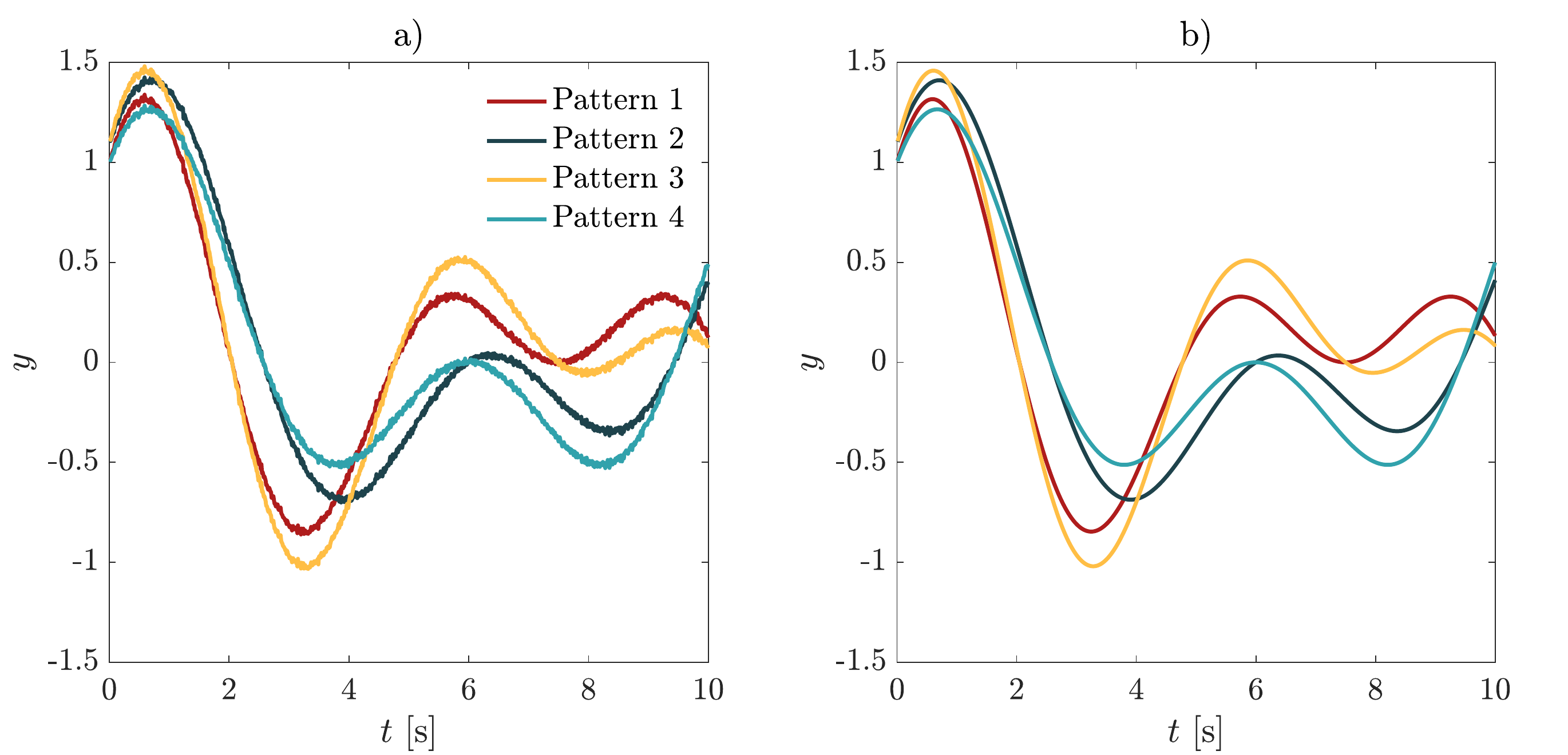}
  \caption{Mean vectors of four patterns of $\bm Y$ (left)
compared with the deterministic part of Eq.\eqref{ModFunEx1} (right)
.}
  \label{fig:Ex1Fig5}
\end{figure}

\begin{figure}[H]
  \centering
  \includegraphics[scale=2.43]{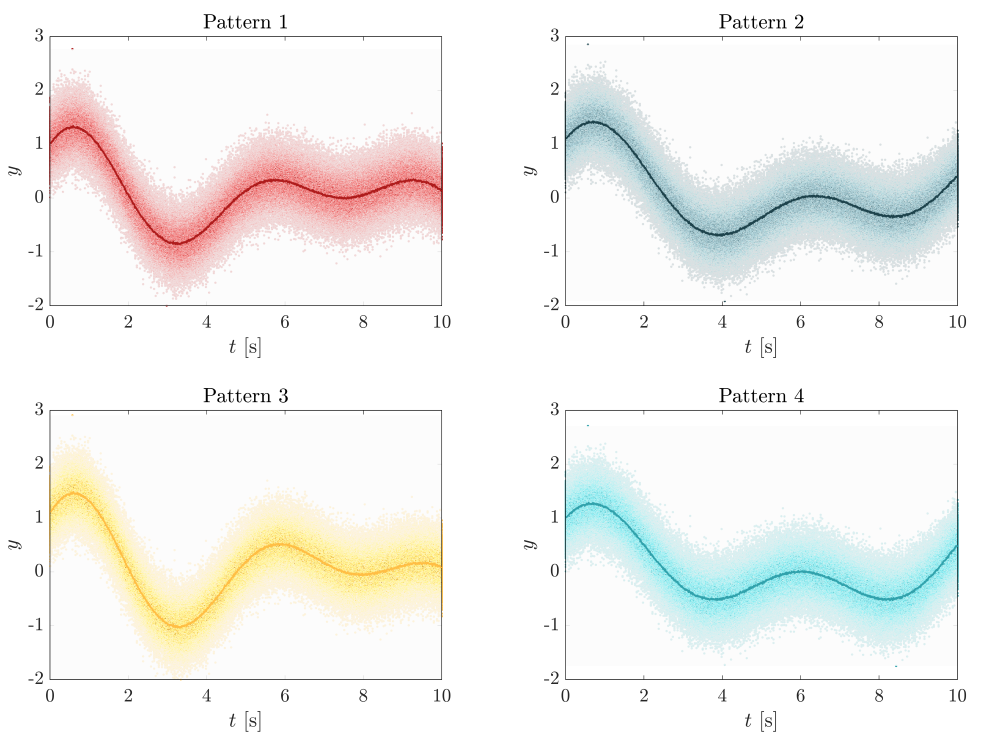}
  \caption{Samples of $\bm Y$ corresponding to each pattern.}
  \label{fig:Ex1Fig6}
\end{figure}

Guided by the feature mapping, Figure \ref{fig:Ex1Fig5} shows the mean vectors of the four performance patterns obtained from a k-means clustering, compared with the deterministic part of Eq.\eqref{ModFunEx1}. Figure \ref{fig:Ex1Fig6} shows the samples of $\bm Y$ corresponding to each performance pattern. It can be seen from Figure \ref{fig:Ex1Fig5} that the mean vectors fully capture the deterministic component of Eq.\eqref{ModFunEx1}. It can also be observed from Figure \ref{fig:Ex1Fig5} and Figure \ref{fig:Ex1Fig6} that Pattern 1 is only slightly different from Pattern 3, and Pattern 2 is only slightly different from Pattern 4, while the difference between Pattern 1/Pattern 3 and Pattern 2/Pattern 4 is significant. This observation is in accordance with the conclusion implied from multiple time-scale diffusion maps. Trivially, the participation factors of each pattern are found to be around $1/4$. 
If parametric descriptions on performance patterns and generating densities are of interest, one could obtain a parametrization of $f_{\bm Y}(\bm y)$ by a Gaussian mixture model with four components. Clearly, the mean vectors of the Gaussian components can be set to the vectors in Figure \ref{fig:Ex1Fig5}, the covariance matrices are close to identity matrices, and each component weight is close to $1/4$. The generating densities can be parameterized by

\begin{equation}
f_{\bm X}(\bm x)=f_{\bm X}(\bm x_p,x_{st})=\sum_{k=1}^{4}\lambda_k \delta(x_{st}–k)f_{\bm X}(\bm x_p|x_{st}=k)\,,\label{GDensityEx1}
\end{equation}
where $f_{\bm X}(\bm x_p|x_{st}=k)$ can be parameterized by Gaussian distributions. The Dirac function appears in Eq.\eqref{GDensityEx1} because there is a discrete random variable.

Finally, it is of interest to consider the case that $X_{st}$ cannot be observed. In this case the generating densities can only be defined in the space of $\bm X_p$. In this case the generating densities parameterized by a Gaussian mixture model are devoid of identifiability, i.e. each Gaussian component in the mixture cannot be differentiated from the others. This is because $\bm X_p$ merely adds random noises to the output (see Eq.\eqref{ModFunEx1}), and the identifiability of $\bm X$ comes from the $X_{st}$ component. However, knowing the fact that the generating densities lack identifiability is a meaningful observation, since this implies there are missing basic random variables or the performance patterns stem from deterministic mechanisms.

\subsection{A stochastic Lorenz system}\label{Lorenz}
\noindent
Consider a Lorenz system described by the following ordinary differential equations \cite{lorenz1963deterministic},

\begin{equation}
\begin{aligned}
  &\frac{d\,y_1}{d\,t}=\sigma(y_2-y_1)\\
  &\frac{d\,y_2}{d\,t}=y_1(\rho-y_3)-y_2\\
  &\frac{d\,y_3}{d\,t}=y_1y_2-\beta y_3\\
\end{aligned}\label{ModFunEx2}
\end{equation}
where $\sigma$, $\rho$ and $\beta$ are system parameters. Lorenz system was originally developed to model convection rolls in the atmosphere, but it could also be used to describe the motion of certain mechanical systems (e.g., Lorenz Waterwheel \cite{matson2007the}). In this example, we set $\sigma=10$, $\beta=8/3$ and $\rho$ to be a Gaussian random variable with mean 24 and variance 1. The initial condition of Eq.\eqref{ModFunEx2}, $[y_1(0),y_2(0)]$, is set to be a bi-variate Gaussian random variable with zero mean and identity covariance matrix, while $y_3(0)$ is fixed to zero. In the context of PPPD, the basic random variables, $\bm X$, are $X=[\rho,y_1(0),y_2(0)]$, and $\bm X$ is a multivariate Gaussian random variable with mean $[24,0,0]$ and identity covariance matrix. The response variables, $\bm Y$, are discretized random processes describing the time evolution of $[y_1,y_2,y_3]$, i.e., $\bm Y=[\bm y_1,\bm y_2,\bm y_3]$, and $\bm y_j=[y_j(t_1 ),…,y_j(t_n )]$, $j=1,2,3$. It is assumed the whole sample space of $\bm Y$ is of interest. The Lorenz system is simulated from time 0 to 100, with a uniform incremental time step of 0.01. Therefore, the dimension of $\bm Y$ is $3\times(100/0.01+1)=30003$.

\begin{figure}[H]
  \centering
  \includegraphics[scale=1.2]{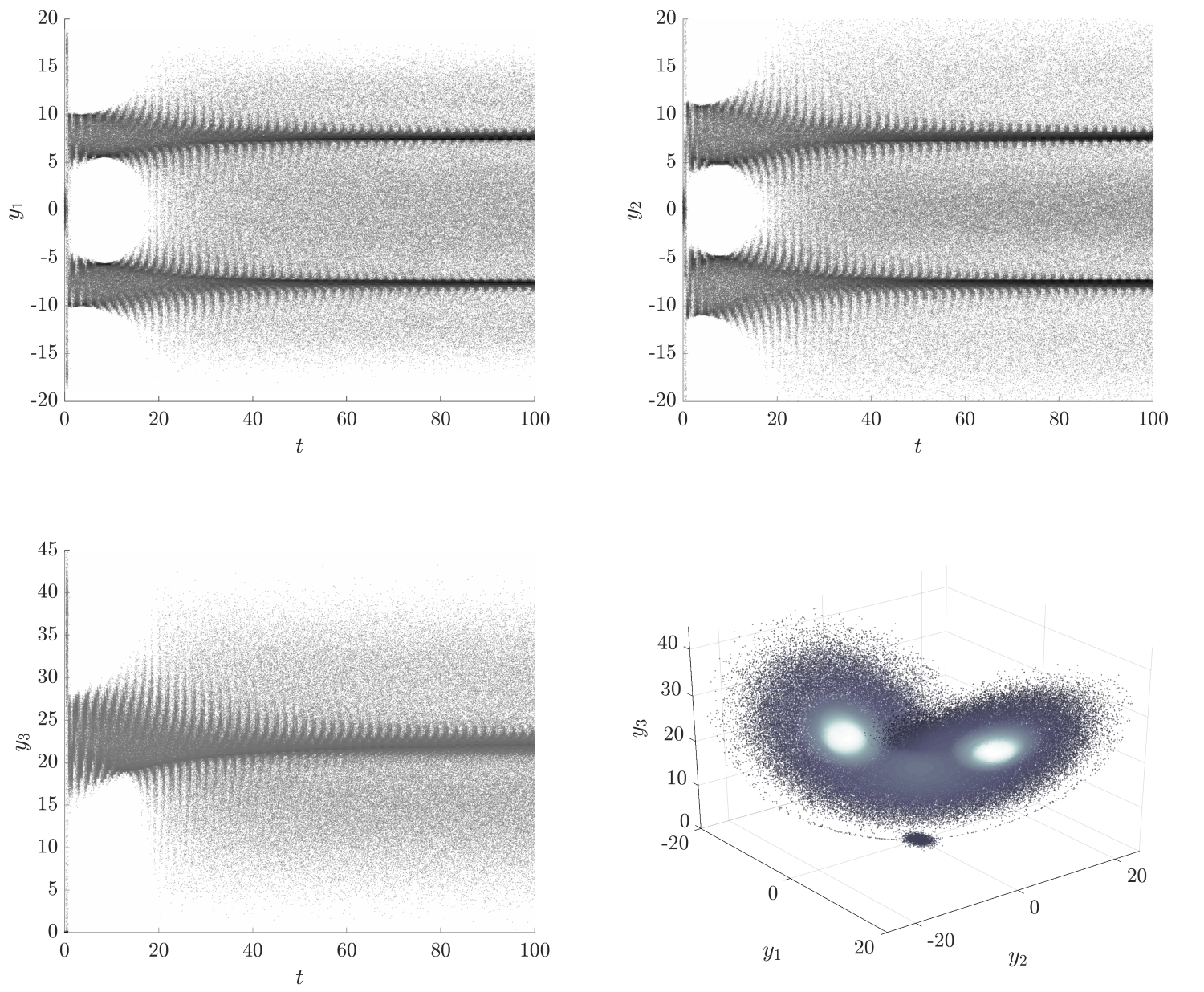}
  \caption{Realizations of $\bm Y$ in the original space.}
  \label{fig:Ex2Fig1}
\end{figure}

Using a direct Monte Carlo simulation and Runge–Kutta method, 2000 random realizations (shown in Figure \ref{fig:Ex2Fig1}) of $\bm Y$ are obtained. Clearly it is difficult to acquire any in-depth understandings on the stochastic Lorenz system by a visual inspection on Figure \ref{fig:Ex2Fig1}.

\begin{figure}[H]
  \centering
  \includegraphics[scale=1.25]{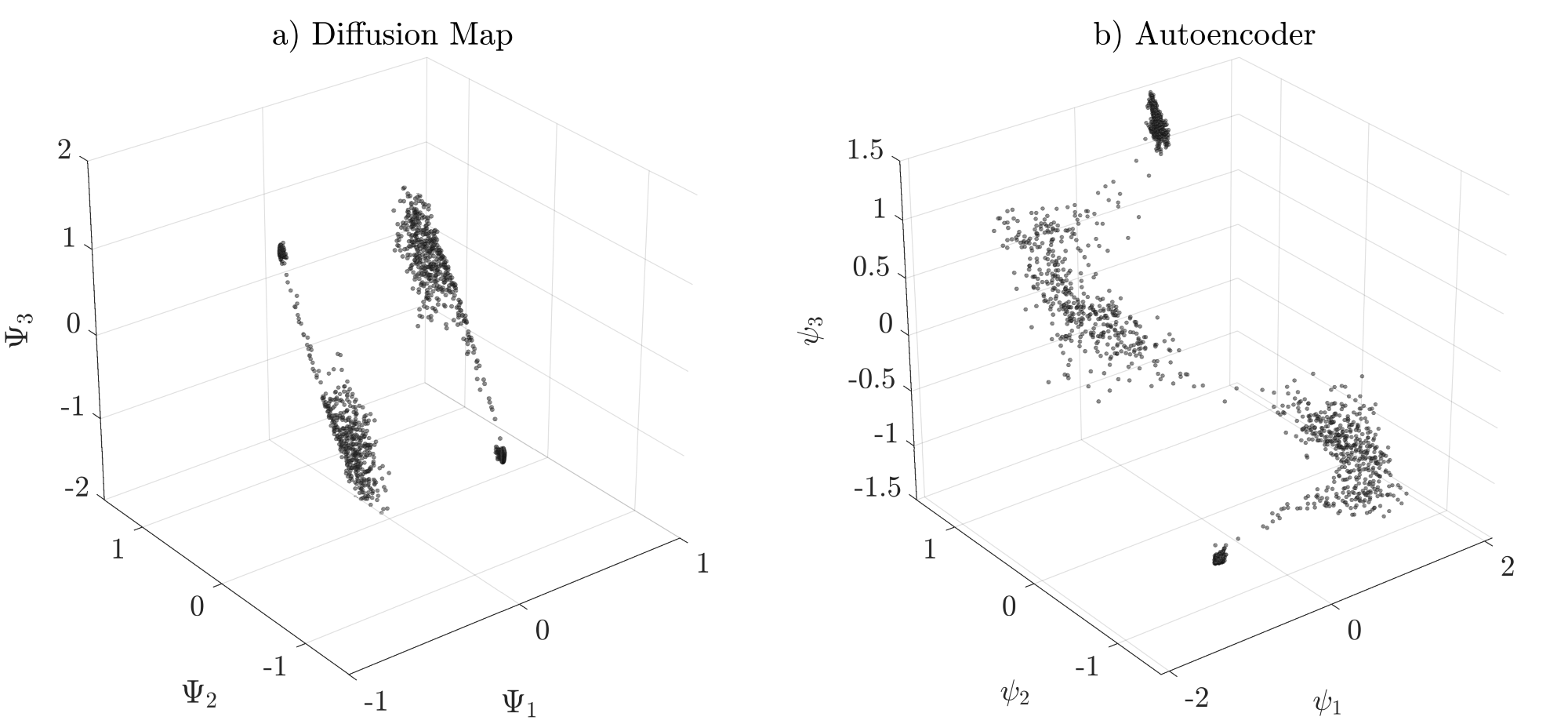}
  \caption{Feature space representation using the diffusion map and the autoencoder.}
  \label{fig:Ex2Fig2}
\end{figure}

Figure \ref{fig:Ex2Fig2} shows realizations of $\bm Y$ embedded into a 3-dimensional feature space, obtained from the diffusion map (with identical settings as that in the previous example), and a 5 hidden layer autoencoder (with identical settings as that in the previous example). For the application of diffusion map in this example, we do not observe qualitatively different behaviors by varying time-scale $\tau$ in a relatively wide range, thus only the diffusion map with $\tau=1$ is illustrated. 

\begin{figure}[H]
  \centering
  \includegraphics[scale=1.75]{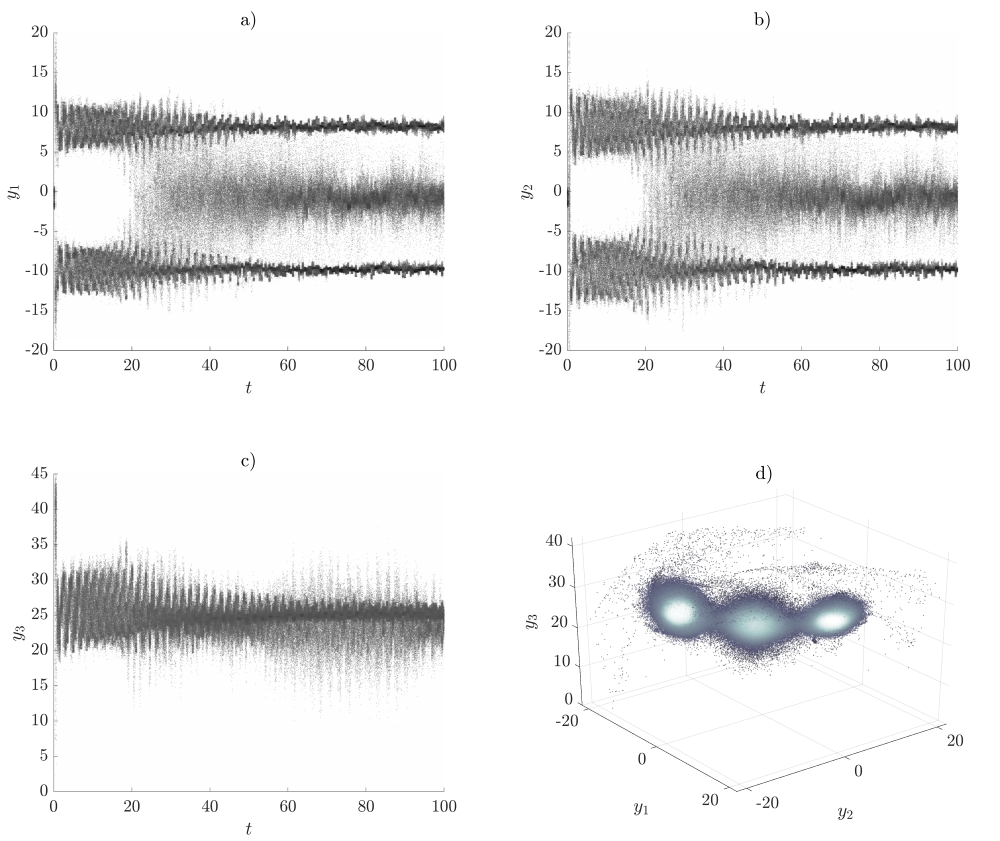}
  \caption{Reconstruction using the autoencoder.}
  \label{fig:Ex2Fig3}
\end{figure}

The reconstructed $\bm Y$ from the autoencoder is shown in Figure \ref{fig:Ex2Fig3}. It can be observed from the feature space representation that: (a) the random trajectories of the Lorenz system can be classified into four patterns; and (b) the four patterns can be further divided into two groups, in one group the samples are tightly clustered while in the other the samples are dispersed. According to properties of Lorenz systems, at this point it is reasonable to conjecture that the aforementioned four patterns are associated with periodic trajectories (where there are two \textit{attractors}) and chaotic trajectories (where there are two \textit{repellors}). Incidentally, one may observe from Figure \ref{fig:Ex2Fig3} that the reconstructed trajectories are in low accuracy. However, in the context of this study, as long as the main features are captured, the reconstruction quality of the autoencoder is not of much practical importance.

\begin{figure}[H]
  \centering
  \includegraphics[scale=1.25]{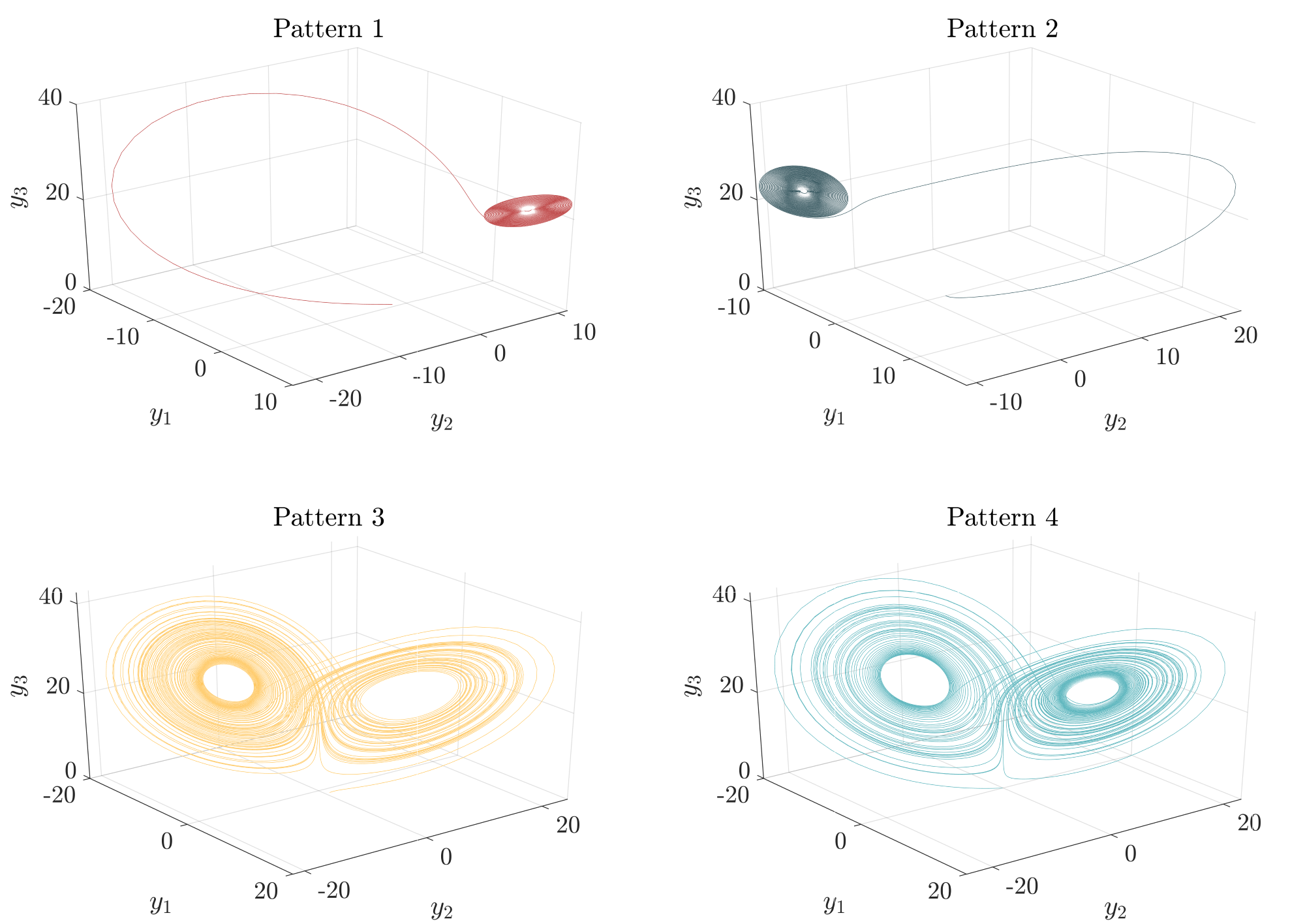}
  \caption{Characteristic trajectories for each pattern.}
  \label{fig:Ex2Fig4}
\end{figure}

\begin{figure}[H]
  \centering
  \includegraphics[scale=1.]{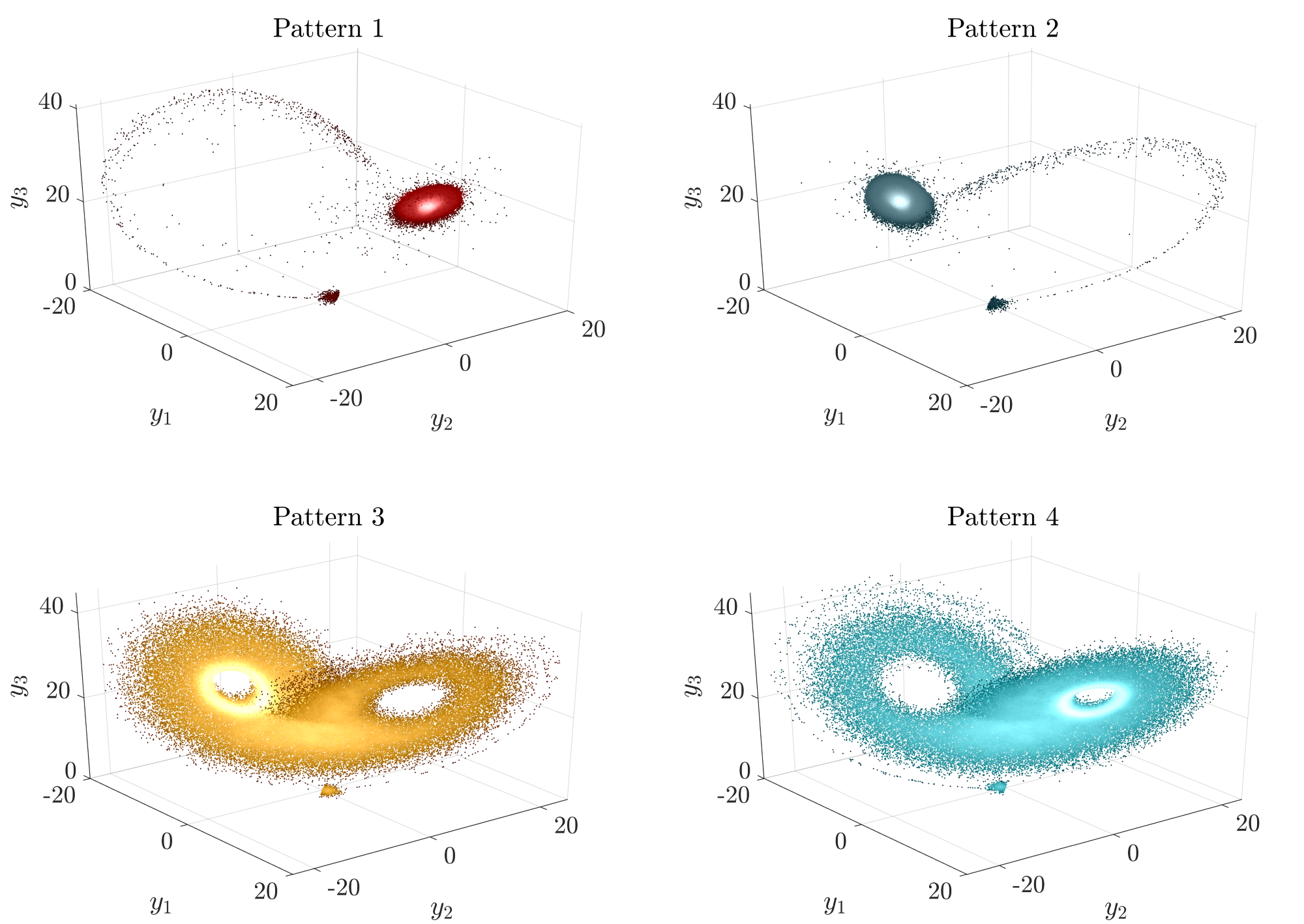}
  \caption{Sample trajectories corresponding to each pattern.}
  \label{fig:Ex2Fig5}
\end{figure}

Figure \ref{fig:Ex2Fig4} shows characteristic trajectories of the four patterns obtained from a ``hierarchical density-based spatial clustering of applications with noise (HDBSCAN)'' \cite{campello2015hierarchical} clustering. Figure \ref{fig:Ex2Fig5} shows the samples of $\bm Y$ corresponding to each pattern. The HDBSCAN instead of the simple k-means clustering is used here since HDBSCAN performs better when handling dataset with varying shapes and densities. The characteristic trajectory for each pattern is obtained as the sample closest to the cluster mean.

It can be observed from Figure \ref{fig:Ex2Fig4} and Figure \ref{fig:Ex2Fig5} that Pattern 1/ Pattern 2 correspond to periodic trajectories in which the system eventually oscillates around one of the two attractors, while Pattern 3/ Pattern 4 correspond to chaotic trajectories in which the system is repelled by the two repellors and exhibit complex behavior. The participation factors of the four patterns are estimated as 0.267, 0.256, 0.243 and 0.234, for Pattern 1, 2, 3 and 4, respectively. 

\begin{figure}[H]
  \centering
  \includegraphics[scale=1.]{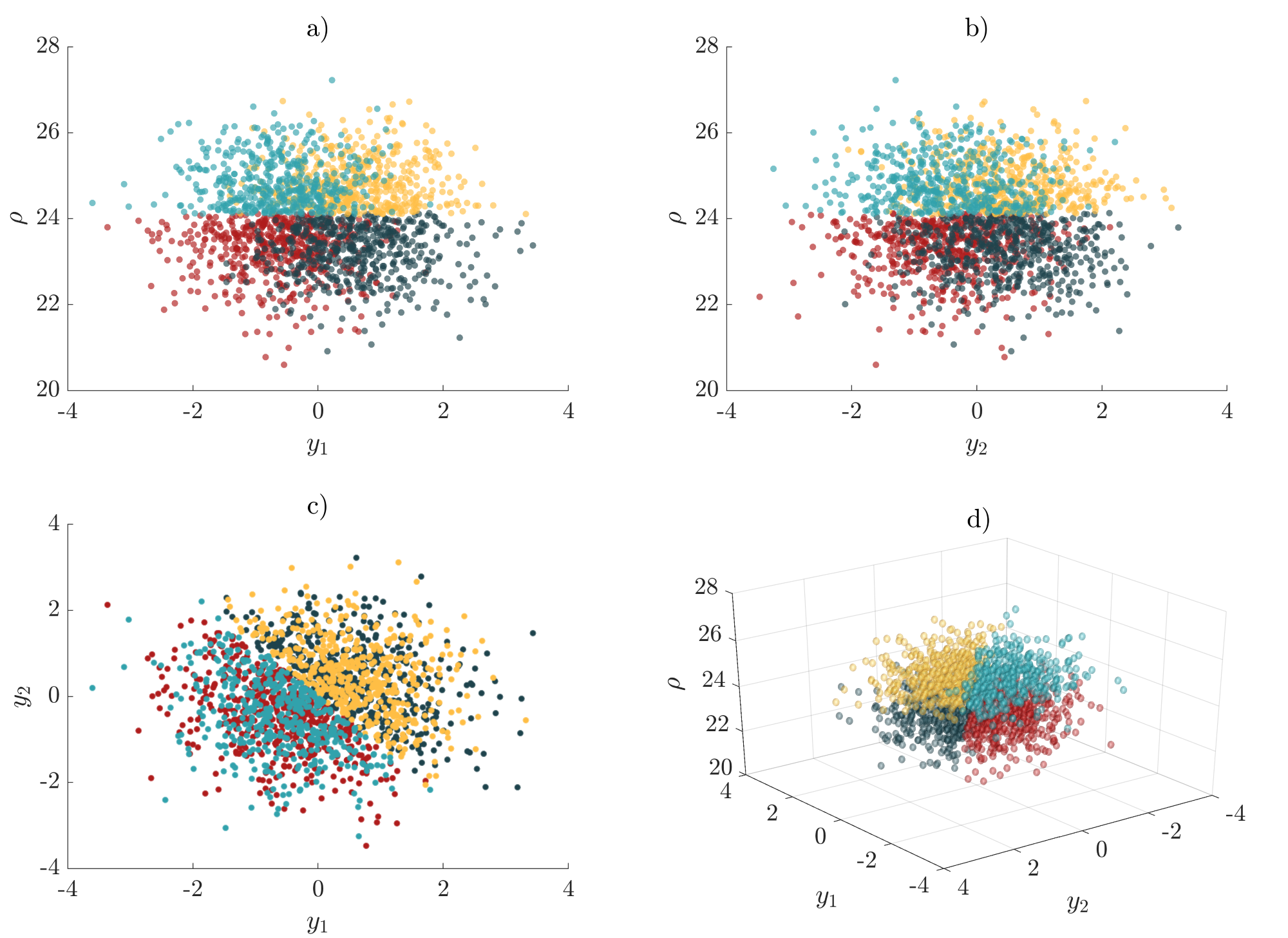}
  \caption{Realizations of basic random variables corresponding to each pattern.}
  \label{fig:Ex2Fig6}
\end{figure}

Next, Figure \ref{fig:Ex2Fig6} illustrates how the patterns are triggered in the sample space of basic random variables, $\bm X=[\rho,y_1(0),y_2(0)]$. One can observe a clear boundary in the $y_1(0)$-$y_2(0)$ plane that separates Pattern 1/Pattern 4 from Pattern 2/Pattern 3. This is because the initial trajectories (trajectories near the initial state) for Pattern 1/Pattern 4 (or Pattern 2/Pattern 3) are similar and they are controlled by the initial condition $[y_1 (0),y_2(0)]$. One can also see that for relatively large $\rho$ values the Lorenz system is chaotic, and for relatively small $\rho$ values the system is periodic. In fact, the smallest $\rho$ value for samples in Pattern 3/Pattern 4 is 24.09, which is fairly close to the theoretical critical $\rho^*=24.06$ \footnote{A critical $\rho$ of 24.06 means that a \textit{strange attractor} corresponds to chaotic trajectories appears at $\rho>24.06$.}.

\subsection{An earthquake engineering example}
\begin{figure}[H]
  \centering
  \includegraphics[scale=0.8]{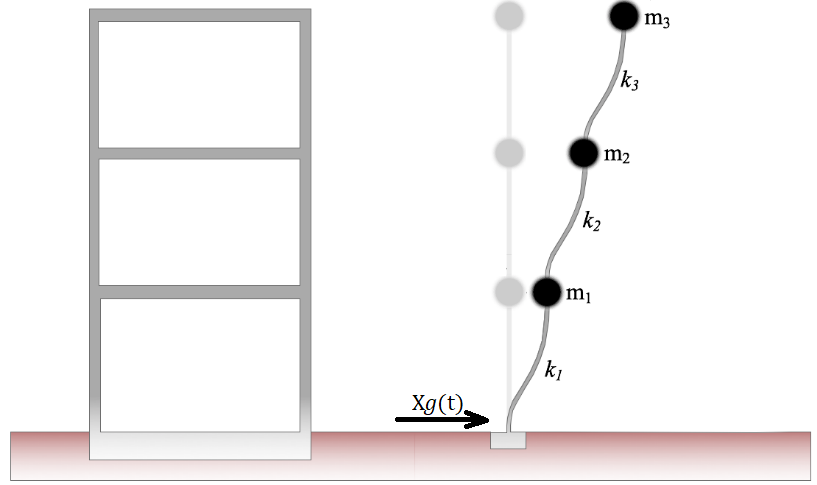}
  \caption{Shear-building model.}
  \label{fig:Ex3Fig1}
\end{figure}

\noindent
Consider a 3-story shear-building model shown in Figure \ref{fig:Ex3Fig1}. The building model is subjected to stochastic ground motion excitation. The force-deformation behavior of each column is assumed to be linearly elastic. The stiffness of each column, $k_1$, $k_2$ and $k_3$, independently follows a log-normal distribution with mean $6.0\times10^7$ [N/m] and coefficient of variation (c.o.v.) of 0.05. The floor masses are identical and equal to $3\times10^4$ [kg], and  5\% damping ratio is assumed for each mode. The building is subjected to a stochastic ground motion with the auto power spectrum density (PSD) described by a modified Kanai-Tajimi model suggested by Clough and Penzien \cite{clough1975st},

\begin{equation}
S_f(\omega)=S_0\frac{\omega_f^4+4\zeta_f^2\omega_f^2\omega^2}{(\omega_f^2-\omega^2)^2+4\zeta_f^2\omega_f^2\omega^2}\frac{1}{(\omega_s^2-\omega^2)^2+4\zeta_s^2\omega_s^2\omega^2}\,,\label{PSD}
\end{equation}
where $S_0=0.0015\rm{[m^2/s^3]}$ is a scale factor, $\omega_f=15$ [rad/s] and $\zeta_f=0.6$ are the filter parameters representing, respectively, the natural frequency and damping ratio of the soil layer, and $\omega_s=0.5$ [rad/s] and $\zeta_s=0.6$ are parameters of a second filter that is introduced to assure finite variance of the ground displacement. The duration of the ground motion is assumed to be 10 seconds.

The stochastic ground motion process $X_g(t)$ is discretized in frequency domain as \cite{timeseries}

\begin{equation}
X_g(t)=\sum_{j=1}^{p/2}\sigma(\omega_j)(x_j\cos(\omega_jt)+x_j'\sin(\omega_jt))\,,\label{GroundMotion}
\end{equation}
where $x_j$, $x_j'$ are independent standard Gaussian variables, the frequency point is given by $\omega_j=j\Delta\omega$ with $p/2=200$, the cut-off frequency is set to $\omega_{p/2}=15\pi$ (therefore $\Delta\omega=30\pi/p=0.075\pi$), and $\sigma(\omega_j)=\sqrt{2S_f(\omega_j)\Delta\omega}$.

Given the specifications of the stochastic process $X_g (t)$, the set of basic random variables, $\bm X$, can be written as $\bm X=[x_1,x_1',…,x_{200},x_{200}',k_1,k_2,k_3 ]$, and the dimension of $\bm X$ is $400+3$ (400 for ground motion and 3 for random stiffness). The response variables $\bm Y$ are discretized random processes describing the time evolution of each inter-story displacement (i.e. relative displacement between roof and ground for each story), and is written as $\bm Y=[\bm y_1,\bm y_2,\bm y_3]$, and $\bm y_j=[y_j (t_1),…,y_j (t_n)]$, $j=1,2,3$. The shear-building model is simulated from time 0 to 10 seconds, with a uniform incremental time step of 0.01. Therefore, the dimension of $\bm Y$ is $3\times(10/0.01+1)=3003$. We are interested in the performance state defined by

\begin{equation}
\mathcal{P}_y=\left\lbrace\bm Y|c-\max\left|\bm Y\right|\leq0\right\rbrace\,,\label{PerfStateEx3}
\end{equation}
where $c$ is a threshold value for the inter-story displacement. 

\begin{figure}[H]
  \centering
  \includegraphics[scale=1.15]{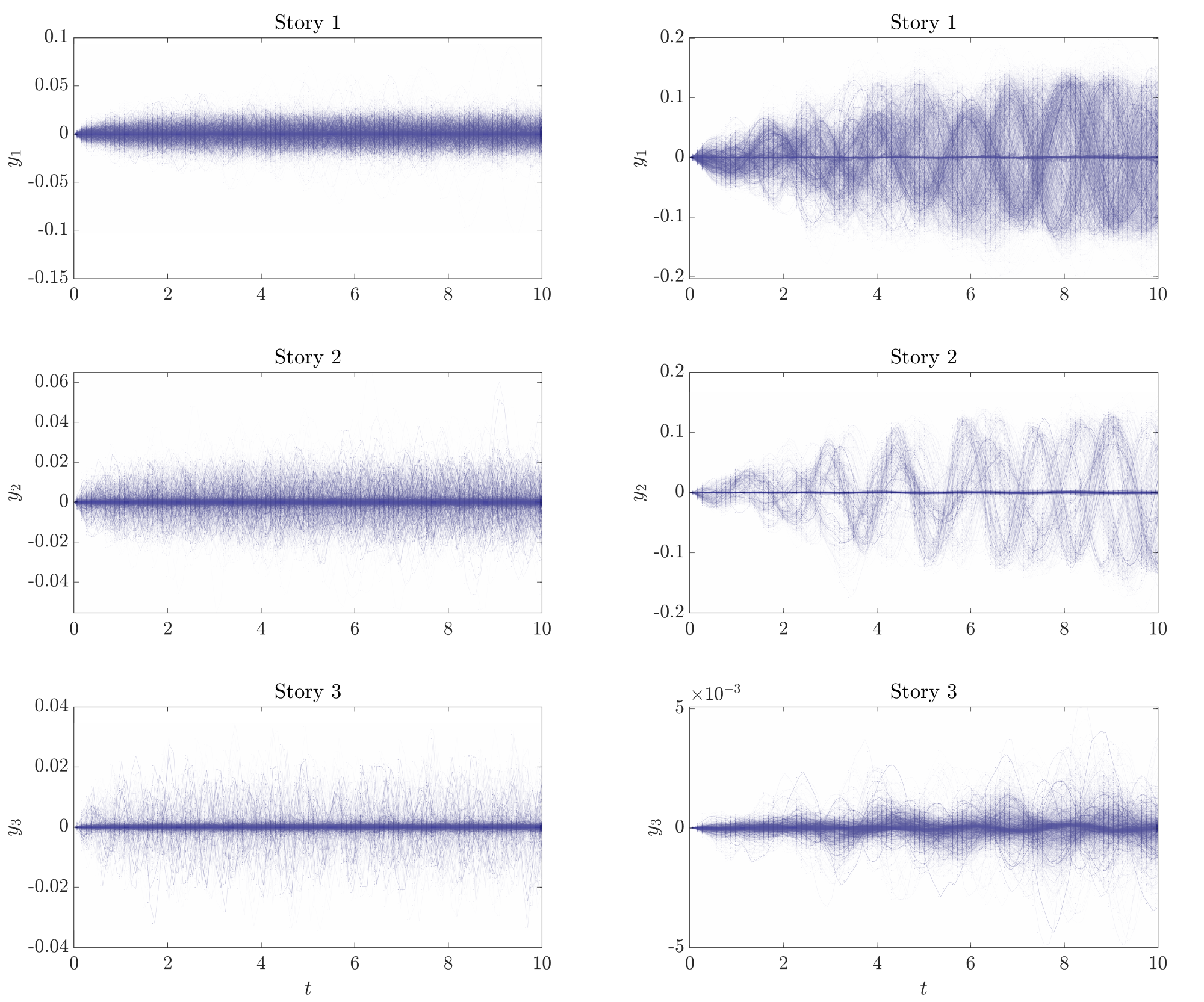}
  \caption{Realizations of $\bm Y$ in the original space for thresholds $c=0.02$ [m] (left) and $c=0.12$ [m] (right).}
  \label{fig:Ex3Fig2}
\end{figure}

Using a Hamiltonian Monte Carlo based sequential Monte Carlo simulation (see \ref{AppendSampling} and \cite{WANG201951}), for threshold values $c=0.02$ [m] and $c=0.12$ [m], we obtain 5000 random realizations (shown in Figure \ref{fig:Ex3Fig2}). The probabilities of $\bm Y\in\mathcal{P}_y$ for threshold values $c=0.02$ [m] and $c=0.12$ [m] are estimated as $3.4\times10^{-2}$ and $1.2\times10^{-7}$, respectively. It is seen from Figure \ref{fig:Ex3Fig2} that for larger threshold, the response of story 3 is surprisingly smaller. This phenomenon can be qualitatively understood as: the system has to find ``efficient'' route to enable any of the inter story displacement to exceed a high response threshold, and it is not efficient for the story 3 to attain a high response value. As the threshold value increases the possibility that story 3 achieves a high response value is ruled out in a natural selection manner. The following discussions in this section will provide further evidence to support the aforementioned idea.

\begin{figure}[H]
  \centering
  \includegraphics[scale=1.04]{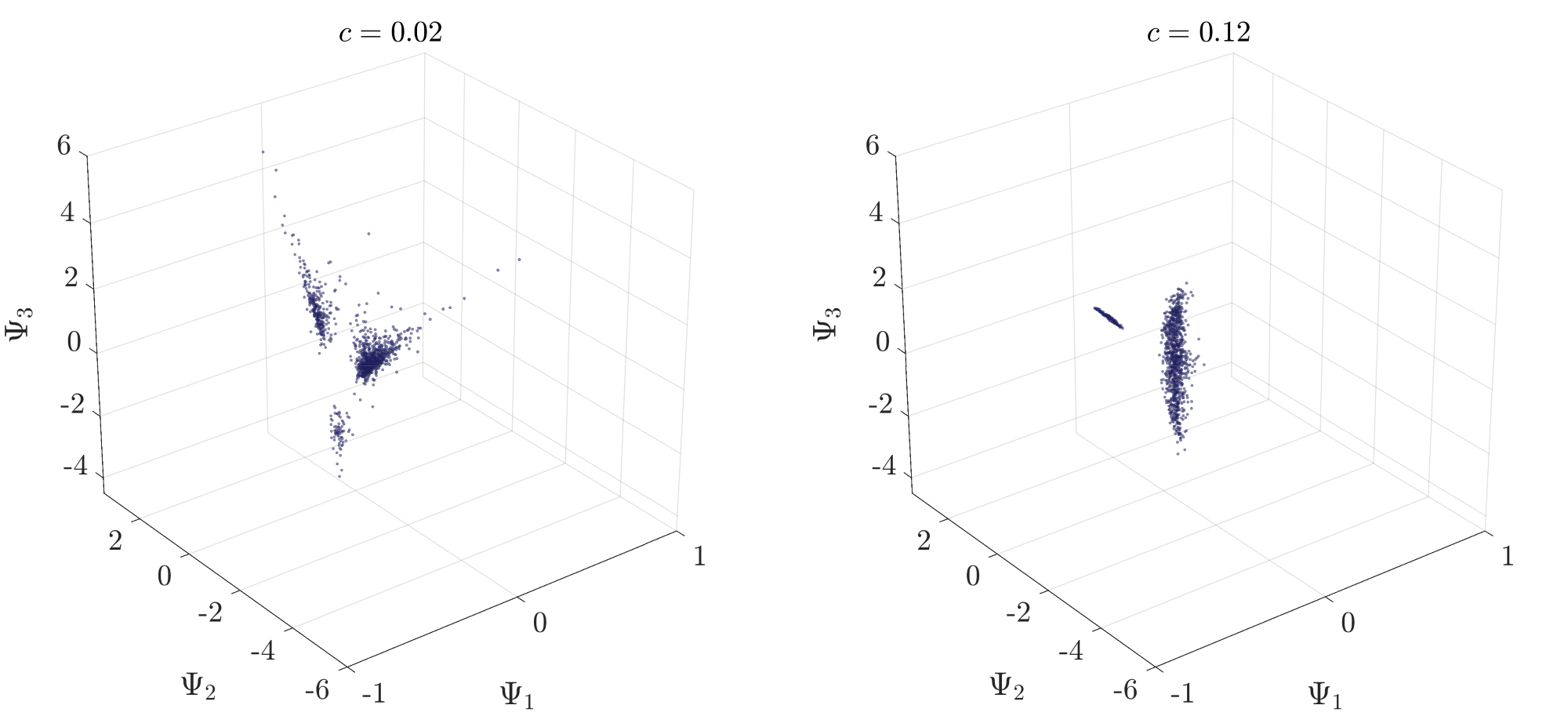}
  \caption{Feature space representation using the diffusion map for thresholds $c=0.02$ [m] (left) and $c=0.12$ [m] (right).}
  \label{fig:Ex3Fig3}
\end{figure}

Figure \ref{fig:Ex3Fig3} shows realizations of $\bm Y$ embedded into a 3-dimensional feature space, obtained from the diffusion map (with identical settings as that in the previous examples). It can be observed from the figure that as the threshold increases, the number of performance patterns changes from 3 to 2.

\begin{figure}[H]
  \centering
  \includegraphics[scale=.58]{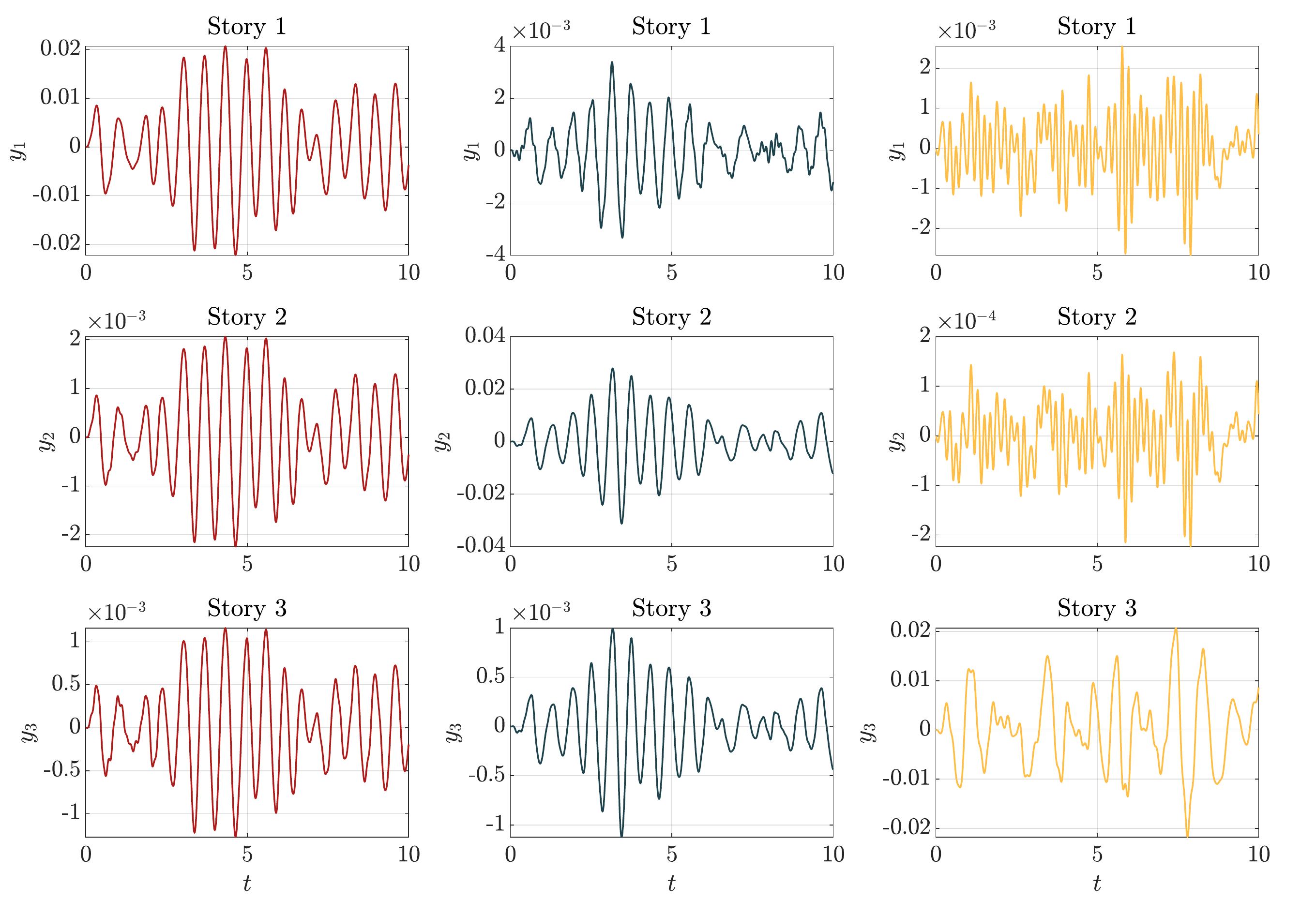}
  \caption{Characteristic trajectories for each pattern (threshold $c=0.02$ [m]). \textit{The left, middle, and right column of plots show the Pattern 1, Pattern 2, and Pattern 3, respectively.}}
  \label{fig:Ex3Fig4}
\end{figure}

\begin{figure}[H]
  \centering
  \includegraphics[scale=.58]{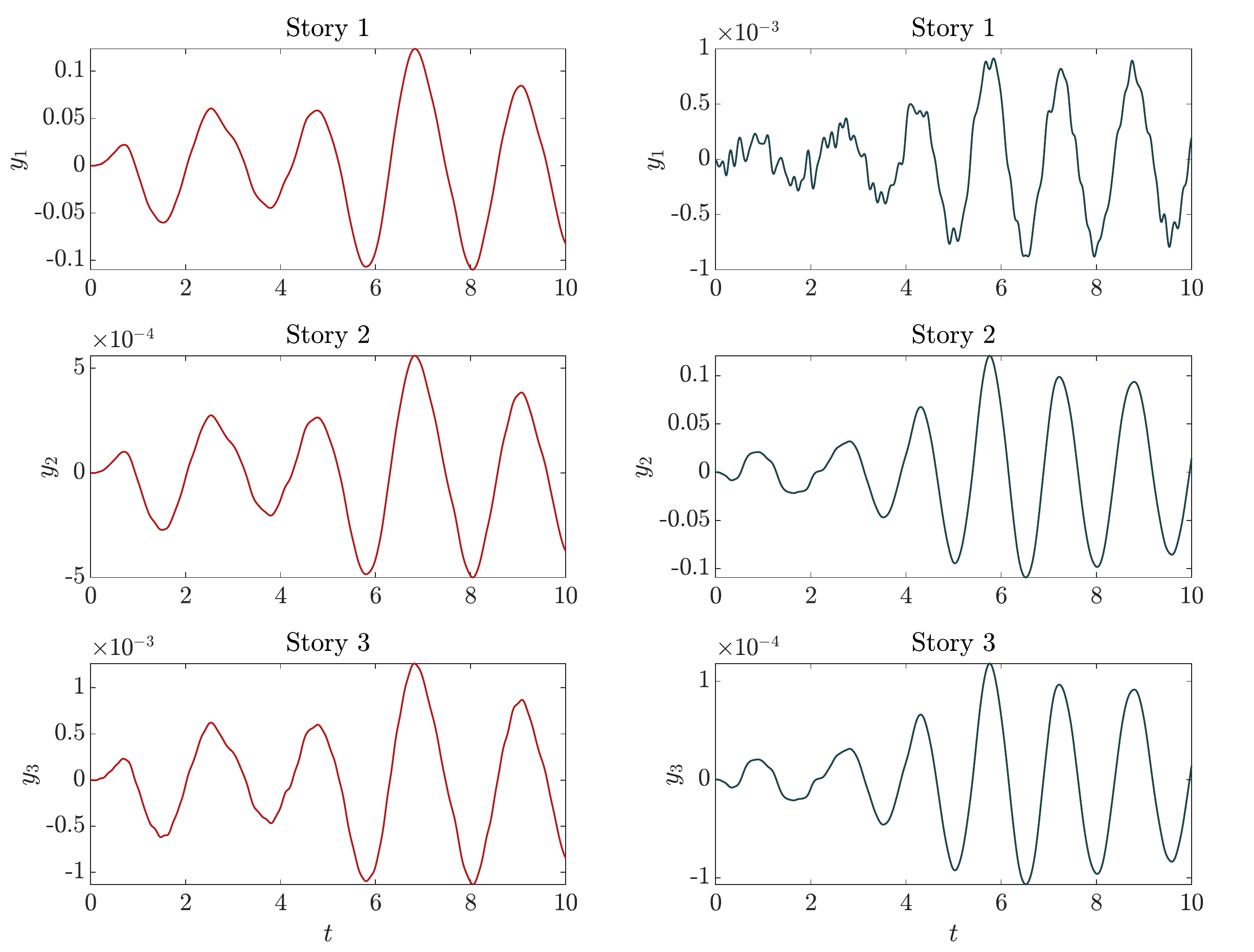}
  \caption{Characteristic trajectories for each pattern (threshold $c=0.12$ [m]). \textit{The left and right column of plots show the Pattern 1 and Pattern 2, respectively.}}
  \label{fig:Ex3Fig5}
\end{figure}

For the two thresholds, pattern identification analysis with k-means clustering is performed. Figure \ref{fig:Ex3Fig4} and Figure \ref{fig:Ex3Fig5} show characteristic trajectories of the performance patterns. The characteristic trajectory for each pattern is obtained as the sample closest to the cluster mean. The participation factors of each pattern for threshold $c=0.02$ [m] are estimated as 0.66, 0.27 and 0.07 for Pattern 1,2 and 3, respectively, while the participation factors for threshold $c=0.12$ [m] are 0.82 and 0.18 for Pattern 1 and 2, respectively.

The following remarks can be made on the performance patterns.

(a) For threshold $c=0.02$ [m], the three patterns correspond to the degree of dominance of each inter story displacement. In Pattern 1 the first inter story displacement is in general larger than the other two stories, in Pattern 2 the second inter story displacement dominates, and in Pattern 3 the third inter story displacement dominates.

(b) For threshold $c=0.02$ [m], the participation factor for each pattern (0.66, 0.27 and 0.07) indicates that it is most likely that the first inter story displacement being larger than the other stories, and it is least likely that the third inter story displacement dominates. This observation is in accordance with common sense (note that the inertia force applied to the first story is the largest, and for this example the mean stiffness of each story is the same).

\begin{figure}[H]
  \centering
  \includegraphics[scale=1.18]{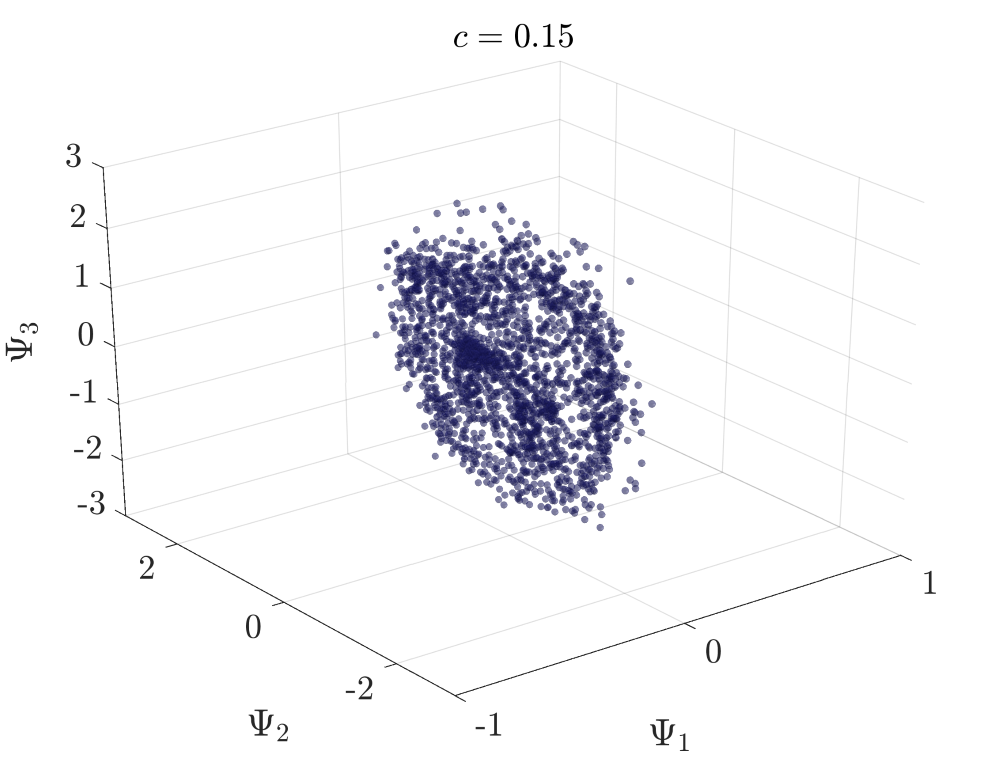}
  \caption{Feature space representation using the diffusion map for threshold $c=0.15$ [m].}
  \label{fig:Ex3Fig6}
\end{figure}

(c) For threshold $c=0.12$ [m], the Pattern 3 in threshold $c=0.02$ [m] disappears\footnote{Rigorously speaking, if not being disappeared, the possibility is extremely small (smaller than $1.2\times10^{-7}\times1/5000\approxeq2.4\times10^{-11}$, recall that we have simulated 5000 events lie in the performance state).}, and the Pattern 1 and Pattern 2 are retained. Given this trend, it is reasonable to conjecture that if the threshold is set even higher, only one pattern (the one corresponds to the first story displacement dominates scenario) would be left. This conjecture is confirmed by performing PPPD analysis for $c=0.15$ [m]. The failure probability for $c=0.15$ [m] is estimated as $9.9\times10^{-9}$, and the feature space representation is given by Figure \ref{fig:Ex3Fig6}. It can be seen from Figure \ref{fig:Ex3Fig6} that all points seem cluster together, suggesting there is only one pattern.

(d) From threshold $c=0.02$ to $c=0.12$, the typical performance pattern trajectories, in general, have a frequency shift to the relatively low frequency side.

\begin{figure}[H]
  \centering
  \includegraphics[scale=.5]{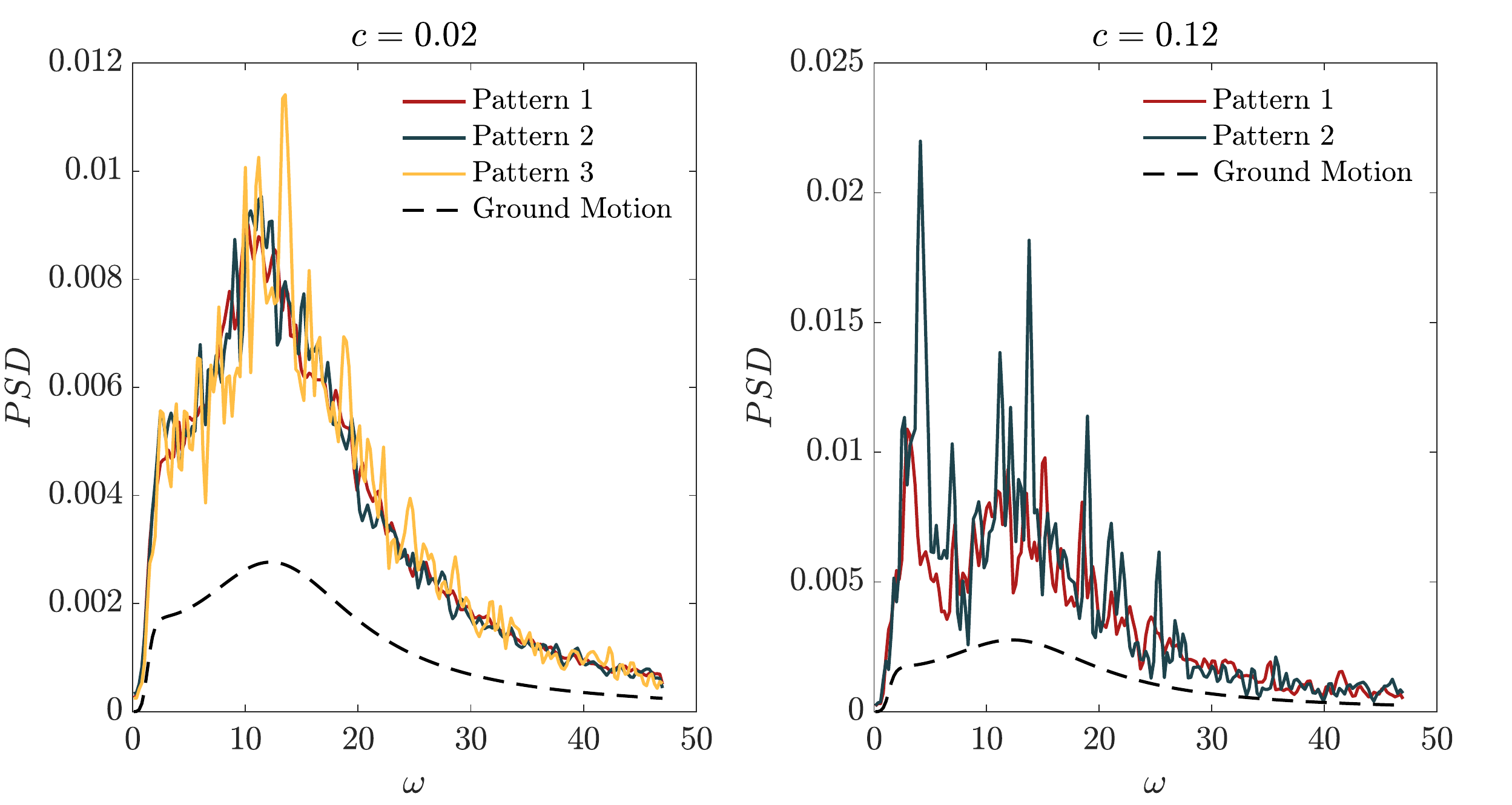}
  \caption{Power spectrum density for ground motion samples of each performance pattern.}
  \label{fig:Ex3Fig7}
\end{figure}

Now we will investigate how the performance patterns are generated in terms of basic random variables. To start with, we investigate if the performance patterns are triggered by ground motions with different characteristics\footnote{Although the power spectrum density model for the stochastic ground motion is fixed (Eq.\eqref{PSD}), since it is a stochastic model the randomly simulated ground motion could still exhibit different characteristics. Therefore, it is possible that the ground motion samples that generate response samples of different performance patterns exhibit different frequency domain characteristics.} (e.g., frequency contents). To have a better illustration, instead of showing the space of $\bm X$ we estimate the power spectrum density (PSD) of ground motion samples\footnote{Recall that the ground motion is a deterministic function of $\bm X$, as shown in Eq.\eqref{GroundMotion}.} corresponding to each pattern, and the results are shown in Figure \ref{fig:Ex3Fig7}. The analytical auto-PSD model of the ground motion (Eq.\eqref{PSD}) is also shown in the figure for a comparison. It is seen from the figure that for each threshold, the PSD curve for each pattern essentially looks similar. Therefore, we conclude that \textbf{for a given threshold}, the performance patterns are not generated by ground motions with different characteristics. However, it is important to observe that this conclusion does not suggest the frequency contents of ground motion do not influence the performance patterns. In fact, it can be observed that the PSD of each performance pattern for threshold $c=0.12$ [m] has richer low frequency contents than that for threshold $c=0.02$ [m].

\begin{figure}[H]
  \centering
  \includegraphics[scale=1.05]{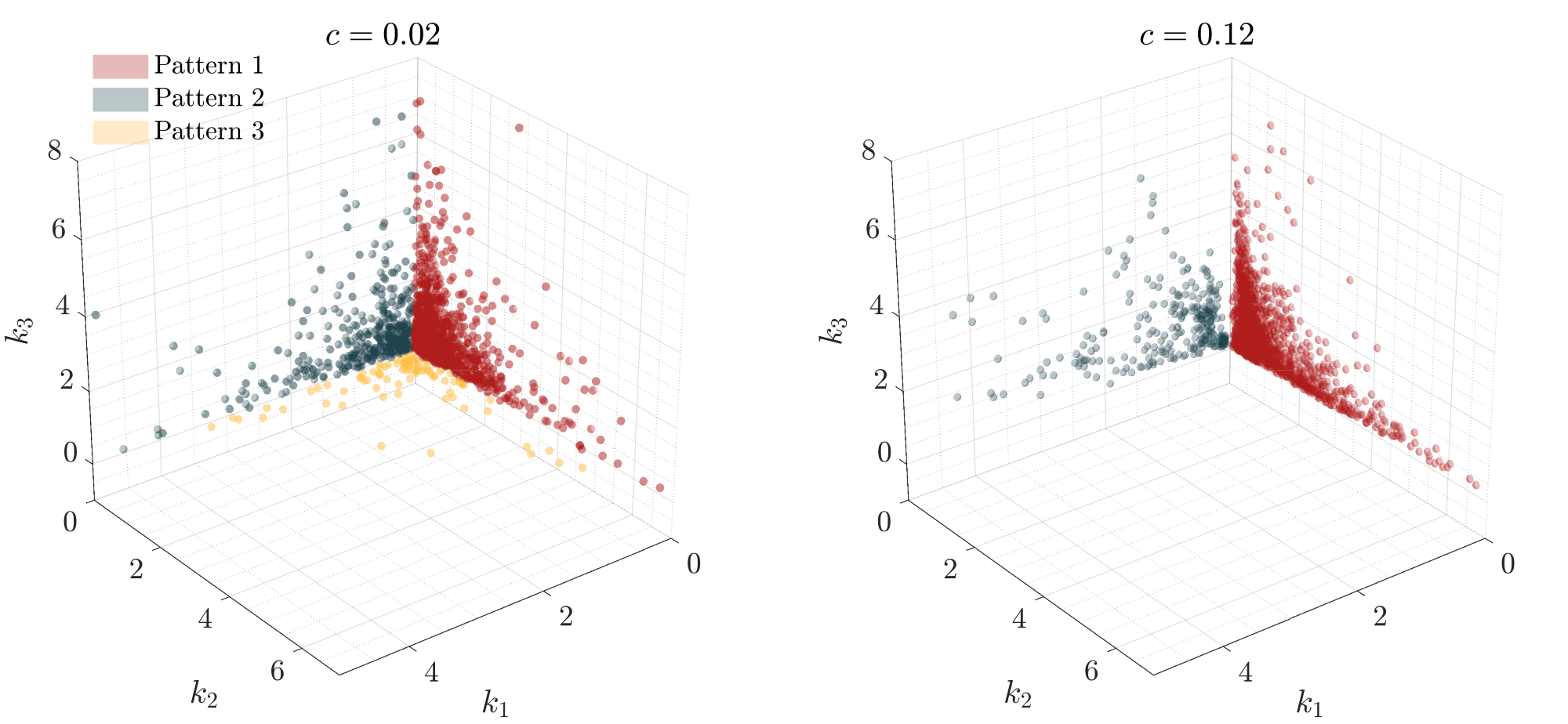}
  \caption{Realizations of random stiffnesses corresponding to each pattern.}
  \label{fig:Ex3Fig8}
\end{figure}

Given the aforementioned investigation, we conjecture that in the space of random stiffnesses $[k_1,k_2,k_3]$, there should be clear patterns. This assumption is confirmed by Figure \ref{fig:Ex3Fig8}, which shows realizations of $[k_1,k_2,k_3]$ corresponding to each performance pattern. Figure \ref{fig:Ex3Fig8} provides a way to design/control the stochastic behavior of the building, so that the random first passage event of maximum responses can be manipulated. Note that in this example Figure \ref{fig:Ex3Fig8} can be qualitatively anticipated, because to have a high likelihood of first passage in certain story, the stiffness at that story should be relatively small. However, PPPD analysis provides the quantitative approach to estimate the most likely setting that triggers certain performance pattern in a rare event.

Finally, it is important to highlight that the results of PPPD are obtained without a knowledge on the underlying physical/mathematical laws that govern the stochastic system. For all the examples studied in this paper, we use the governing laws to generate random samples, however, if the samples are obtained by performing real experiments or collecting data from sensors, the PPPD analysis can be applied in the same manner. This perspective further highlights the potential applications of PPPD analysis.

\section{Conclusions}\label{Conclusion}
\noindent
A framework termed Probabilistic Performance-Pattern Decomposition (PPPD) is developed to facilitate an in-depth understanding on the behaviors of stochastic systems. The concept of performance-pattern is developed using response variables, which directly describe the behavior of a stochastic system, and the performance state, which is a specified subset in the sample space of response variables. The theoretical framework of PPPD is proposed via a probabilistic decomposition of response variables conditional on the performance state. The computational framework of PPPD is consisted of three major ingredients: 1) event observation; 2) feature mapping; and 3) pattern identification. Using rare event simulation, manifold learning and clustering techniques, the computational framework of PPPD is capable of analyzing complex stochastic systems involving random fields/processes, and producing the main behavior patterns of the system conditional on the performance state of interest. Moreover, PPPD analysis enables identify critical domains in the sample space of basic random variables that trigger each performance pattern.

To illustrate the effectiveness of PPPD, the paper investigates three non-trivial numerical examples which all involve random processes and high dimensional probability spaces. The first example is a hypothetical system with analytical stochastic input and output processes. A PPPD analysis for this example results in four performance patterns, which are in close accordance with mathematical rules of the hypothetical system. The second example is a Lorenz system with random system parameters and initial conditions. The PPPD analysis enables one to differentiate between periodic and chaotic response trajectories, and to investigate how different performance patterns can be generated. The last example is a simplified shear-building model with random stiffnesses and subjected to a stochastic ground motion excitation described by a power spectrum density model. A PPPD analysis for this example leads to insightful results on how the performance patterns shift with the decrease of failure probability, and how the patterns are generated in the space of basic random variables.

A promising application of PPPD is to use it in a fully data-driven fashion to discover patterns and regularities of large-scale sophisticated stochastic systems. Ultimately, PPPD can be used to assist physics-informed decision process.

\section*{Acknowledgement}
\noindent
Dr. Ziqi Wang was supported by the National Science and Technology Major Project of the Ministry of Science and Technology of China (Grant No. 2016YFB0200605), the National Natural Science Foundation of China (Grant No. 51808149) and the Provincial Natural Science Foundation of Guangdong Province (Grant No. 2018A030310067). Dr. Marco Broccardo was supported by the Swiss Seismological Service (SED) and by the Chair of Structural Dynamics and Earthquake Engineering at ETH Z\"urich. Prof. Junho Song was supported by the Institute of Construction and Environmental Engineering at Seoul National University, and the project ``Development of Lifecycle Engineering Technique and Construction Method for Global Competitiveness Upgrade of Cable Bridges'' funded by the Ministry of Land, Infrastructure and Transport (MOLIT) of the Korean Government (Grant No. 16SCIP-B119960-01). This support is gratefully acknowledged. Any opinions, findings, and conclusions expressed in this paper are those of the authors, and do not necessarily reflect the views of the sponsors.

\appendix
\section{Sequential Monte Carlo method to sample from the performance state domain}\label{AppendSampling}

\begin{algorithm}[H]
\caption{Sequential Monte Carlo simulation to generate $N$ random realizations from $f_{\bm X}(\bm x|\mathcal{P}_x)$}\label{alg:SMC}
\begin{description}

\item [Step 1: Parameter specification]
\rule{0pt}{15pt}
\begin{itemize}
\item Define $p_0$, the conditional probability for each intermediate states.
\item Define $N_0$, the sample size in each intermediate step. We let $N_0\cdot p_0\approx N$.
\end{itemize}

\item [Step 2: Initial run]
\rule{0pt}{15pt}
\begin{itemize}
\item Draw $N_0$ samples, $\bm{x}_0^{(i)}$, $i=1,2,...,N_0$, from PDF $f_{\bm X}(\bm x)$.
\item Evaluate $\bm{y}_0^{(i)}=\mathcal{M}(\bm{x}_0^{(i)})$ and $G(\bm{y}_0^{(i)})$, $i=1,2,...,N_0$.
\item Sort samples $\bm{x}_0^{(i)}$ and $\bm{y}_0^{(i)}$ in increasing orders of $G(\bm{y}_0^{(i)})$.
\item Find $g^{(1)}$ as the $p_0$ percentile of $G(\bm{y}_0^{(i)})$, so that $\mathcal{P}_x^{(1)}$ is specified as $\mathcal{P}_x^{(1)}=\left\lbrace\bm x|G(\mathcal{M}(\bm x))-g^{(1)}\leq0\right\rbrace$.
\item Set $j\leftarrow1$.
\end{itemize}

\item [Step 3: Iterative runs]
\rule{0pt}{15pt}
\begin{itemize}
\item Repeat while $g^{(j)}>0$
\begin{itemize}
\item[$\blacktriangleright$] Starting from $p_0\cdot N_0$ seed samples $\bm x_{j–1}^{(i)}$, $i=1,...,p_0\cdot N_0$, that have $\bm x_{j–1}^{(i)}\in\mathcal{P}_x^{(j)}$, use a MCMC sampler to drawn $(1–p_0)N_0$ samples from PDF $f_{\bm X}(\bm x|\mathcal{P}_x^{(j)})$.

\item[$\blacktriangleright$] Store the  $p_0\cdot N_0+(1–p_0)N_0=N_0$ samples that lie in $\mathcal{P}_x^{(j)}$ as $\bm x_j^{(i)}$.

\item[$\blacktriangleright$] Sort samples $\bm x_j^{(i)}$ in increasing orders of $G(\bm y_j^{(i)})$, where $\bm y_j^{(i)}=\mathcal{M}(\bm x_j^{(i)})$.

\item[$\blacktriangleright$] Find $g^{(j+1)}$ as the $p_0$ percentile of $G(\bm y_j^{(i)})$, so that $\mathcal{P}_x^{(j+1)}=\left\lbrace\bm x|G(\mathcal{M}(\bm x))-g^{(j+1)}\leq0\right\rbrace$.

\item[$\blacktriangleright$] Set $j\leftarrow j+1$.
\end{itemize}
\end{itemize}

\item [Step 4: Final MCMC sampling]
\rule{0pt}{15pt}
\begin{itemize}
\item Use all samples in $\mathcal{P}_x$ as seeds, perform MCMC sampling until a total of $N$ samples in $\mathcal{P}_x$ are obtained.
\end{itemize}
\end{description}
\end{algorithm}

To have a highly representative set of realizations of $f_{\bm X}(\bm x|\mathcal{P}_x)$ to facilitate PPPD, the MCMC algorithm used in Algorithm \ref{alg:SMC} should be able to effectively explore the performance state. One attractive MCMC algorithm proven to be highly effective in various statistical computing applications is the Hamiltonian Monte Carlo (HMC) method \cite{neal2011mcmc}. Implementation details of the HMC algorithm in the context of SMC can be found in \cite{WANG201951}.

\section{Implementation of diffusion map for PPPD}\label{AppendDMap}
\noindent
For the dataset $\mathcal{\bm Y}$, the basic procedures of constructing feature vectors $\bm\Psi$ using the diffusion map is described as follows. 	

\begin{algorithm}[H]
\caption{Constructing feature vectors $\bm\Psi$ from $\mathcal{\bm Y}$ using the diffusion map}\label{alg:DMap}
\begin{description}

\item [Step 1: Construct the similarity matrix]
\rule{0pt}{15pt}
\begin{itemize}
\item Construct the similarity matrix $\bm W=\left\lbrace w_{ij}\right\rbrace$, $i,j=1,2,...,N$, where $w_{ij}=s(\bm y^{(i)},\bm y^{(j)})$, $s(\cdot)$ is a specified similarity function.
\end{itemize}

\item [Step 2: Obtain the Markov matrix]
\rule{0pt}{15pt}
\begin{itemize}
\item Normalize $\bm W$ by $\widehat{\bm{W}}=\bm D^{–\alpha}\bm W\bm D^{–\alpha}$, where $\bm D$ is a diagonal matrix with $\bm D_{ii}=\sum_{j=1}^{N}w_{ij}$, and $\alpha$, $\alpha\in\mathbb{R}$, is a specified parameter.

\item Compute the Markov matrix $\bm M$ by $\bm M=\widehat{\bm D}^{–1}\widehat{\bm{W}}$, where $\widehat{\bm D}$ is a diagonal matrix with $\widehat{\bm D}_{ii}=\sum_{j=1}^{N}\hat{w}_{ij}$.
\end{itemize}

\item [Step 3: Obtain the feature vectors]
\rule{0pt}{15pt}
\begin{itemize}
\item Compute the $n_{t}$ largest eigenvalues and the corresponding eigenvectors for matrix $\bm M$, denoted by $\lambda_i$ and $\phi_i$, $i=1,2,...,n_{t}$, respectively.

\item Compute feature vectors $\bm\Psi=[\bm\psi^{(1)},..,\bm\psi^{(N)}]$ by $\bm\Psi=\bm\Lambda^\tau\bm\Phi^T$, where $\Lambda$ is a $n_t\times n_t$ diagonal matrix with $D_{ii}=\lambda_i$; $\tau$, $\tau\in\mathbb{N}^+$, is a scale parameter describing the time-scale of the diffusion process; $\bm\Phi^T$ is the transpose of the $N\times n_t$ eigenmatrix $\bm\Phi=[\bm\phi_1,…,\bm\phi_{n_t}]$.
\end{itemize}
\end{description}
\end{algorithm}

\subsubsection{Remark 1 of Algorithm \ref{alg:DMap}: selecting the similarity function}
\noindent
The similarity function $s(\cdot)$ could be problem-specific, but has to satisfy: (a) $s(\cdot)$ is symmetric, i.e. $s(\bm y^{(i)},\bm y^{(j)})=s(\bm y^{(j)},\bm y^{(i)})$, and (b) $s(\cdot)$ is positivity preserving, i.e. $s(\cdot)\geq0$. A common choice of $s(\cdot)$ is of the exponential kernel form written as

\begin{equation}
s(\bm y^{(i)},\bm y^{(j)})=\exp\left[-\frac{d^2(\bm y^{(i)},\bm y^{(j)})}{\epsilon}\right]\,,\label{SimilarityM}
\end{equation}
where $d(\cdot)$ is a specified distance function, and $\epsilon$ is a specified scale parameter. In case $d(\cdot)$ is the Mahalanobis distance, Eq.\eqref{SimilarityM} is equivalent to the classical Gaussian kernel.

In PPPD, if one has physical insight or other problem specific intuition on how radically different one realization of response variables is from another, it should be reflected in the definition of $d(\cdot)$. Otherwise, one may use conventional distance measures such as the $L^p$-norm. Note that if an $L^p$-norm distance is used in Eq.\eqref{SimilarityM}, the similarity matrix $\bm W$ will be dense, since every entry of $\bm W$ is nonzero in principle. This would lead to storage and efficiency issues for a large dataset. Clearly, a simple remedy to this dense matrix issue is to convert entries of $\bm W$ with values below some threshold to zero. An alternative approach to obtain a sparse $\bm W$ is to use methods such as k-nearest neighbor (k-NN) \cite{hall2008choice} to determine $s(\bm y^{(i)},\bm y^{(j)})$. Specifically, $s(\bm y^{(i)},\bm y^{(j)})$ can be obtained via: if $\bm y^{(i)}$ is within the k-nearest neighbors of $\bm y^{(j)}$, \textit{or} $\bm y^{(j)}$ is within the k-nearest neighbors of $\bm y^{(i)}$, where ‘k-nearest’ is measured by $d(\cdot)$, then $s(\bm y^{(i)},\bm y^{(j)})$ is computed from Eq.\eqref{SimilarityM}; otherwise, set $s(\bm y^{(i)},\bm y^{(j)})$ to zero. A simple rule of thumb \cite{hall2008choice} to select the parameter ``k'' in k-NN algorithm is to set it in the order of $\log N$, where $N$ is the number of samples in the dataset.

\subsubsection{Remark 2 of Algorithm \ref{alg:DMap}: selecting parameter $\alpha$}
\noindent
The parameter $\alpha$ used in \textbf{Step 2} of the algorithm alters the amount of influence of sample densities over the eigen-functions and spectrum of the diffusion. It is analyzed in \cite{Coifman2006Diffusion} that the parameter settings $\alpha=0$, $\alpha=0.5$, and $\alpha=1$ are particularly meaningful. Specifically, $\alpha=0$ corresponds to a Markov matrix that is identical to the random walk normalized Laplacian in graph theory, and the influence of the sample density is maximal; $\alpha=0.5$ (approximately) corresponds to the diffusion of a Fokker-Planck equation; $\alpha=1$ (approximately) corresponds to the Laplace-Beltrami operator (Brownian motion on the manifold where the data is sampled from), where one is able to recover the Riemannian geometry of the dataset. For the purpose of manifold learning, the setting of $\alpha=1$ is suggested in many applications.

\subsubsection{Remark 3 of Algorithm \ref{alg:DMap}: selecting parameter $\tau$}
\noindent
The parameter $\tau$ corresponds to the number of steps of running the Markov chain (characterized by the Markov matrix $\bm M$) forward in time. Therefore, instead of fixing $\tau$, one could run Algorithm \ref{alg:DMap} for different time-scales to study the multiscale geometry of the dataset.

\section{Implementation of autoencoder for PPPD}\label{AppendAutoEncoder}
\noindent
For the dataset $\mathcal{\bm Y}$, the basic procedures of constructing feature vectors $\bm\Psi$ using the autoencoder is described as follows. 	

\begin{algorithm}[H]
\caption{Constructing feature vectors $\bm\Psi$ from $\mathcal{\bm Y}$ using the autoencoder}\label{alg:AutoEncode}
\begin{description}

\item [Step 1: Define architecture and parameters of the autoencoder]
\rule{0pt}{15pt}
\begin{itemize}
\item Define the number of layers in the encoder/decoder, denoted as $k_{cod}$.
\item Define the number of neurons in each layer of the encoder/decoder, denoted as $n_j$, $j=1,...,k_{cod}$.
\item Define the activation functions for neurons in the encoder and decoder.
\item Define the cost function for training the autoencoder.
\end{itemize}

\item [Step 2: Perform layer-by-layer training]
\rule{0pt}{15pt}
\begin{itemize}
\item Set dataset $\mathcal{\bm D}\leftarrow\mathcal{\bm Y}$.

\item Set $j\leftarrow1$.

\item Repeat while $j\leq k_{code}$
\begin{itemize}
  \item[$\blacktriangleright$] Set $n_{train}\leftarrow n_j$.
  \item[$\blacktriangleright$] Using $\mathcal{\bm D}$ as input, train an autoencoder with a single hidden layer of $n_{train}$ neurons.
  \item[$\blacktriangleright$] Set $\bm\Psi$ as the output of the hidden layer.
  \item[$\blacktriangleright$] Set $\mathcal{\bm D}\leftarrow\bm\Psi$.
  \item[$\blacktriangleright$] Set $j\leftarrow j+1$.
\end{itemize}
\end{itemize}

\item [Step 3: Fine-tuning the whole autoencoder (optional)]
\rule{0pt}{15pt}
\begin{itemize}
\item Stack the single hidden layer autoencoders obtained in \textbf{Step 2} to form the whole deep autoencoder.

\item Perform global fine-tuning for the whole autoencoder to optimize the reconstruction of $\mathcal{\bm Y}$.

\item Using $\mathcal{\bm Y}$ as input for the tuned autoencoder, set $\bm\Psi$ as the output of the bottle-neck layer.
\end{itemize}
\end{description}
\end{algorithm}

It is common practice to set the number of neurons in each hidden layer of the autoencoder to be smaller than the dimension of the input, otherwise there is a risk to learn the identity function. Besides manipulating the network architecture, an alternative approach to enforce the autoencoder learning useful structure is to introduce a sparsity regularization term to the cost function. In general, the cost function for an autoencoder can be of the form \cite{olshausen1997sparse}

\begin{equation}
cost=\frac{1}{N}\sum_{i=1}^{N}d(\bm y^{(i)},\hat{\bm y}^{(i)})+\alpha_{sp}\sum_{j=1}^{n_h}d_{sp}(\rho^{(j)},\bar{\rho}^{(j)})+c_{reg}\,,\label{CostF}
\end{equation}
where $d(\cdot)$ is a specified distance function (identical to that in Eq.\eqref{SimilarityM}); $\rho^{(j)}$ is a specified target activation value for each neuron in the hidden layers and $\bar{\rho}^{(j)}$ is the average activation value (averaged over all samples in the training dataset); $d_{sp}(\cdot)$ is a specified distance function (for sigmoid neurons $d_{sp}(\cdot)$ can be the Kullback-Leibler divergence); $n_h=–n_{k_{cod} }+2\sum_{j=1}^{k_{cod}}n_j$ is the total number of neurons in the hidden layers; $\alpha_{sp}$ is a parameter controls the influence of the sparsity regularization term; $c_{reg}$ denotes other regularization term that might be used (e.g., $L_1$ or $L_2$ regularization).

\bibliography{PPPD}

\end{document}